\pdfoutput=1
\documentclass[runningheads]{llncs}
\usepackage[T1]{fontenc}

\usepackage[square, numbers]{natbib}
\usepackage{booktabs}

\usepackage{amsmath,amssymb,amsfonts}
\usepackage{subcaption}
\usepackage{longtable}
\usepackage{algorithm}
\usepackage[noend]{algpseudocode}
\usepackage{graphicx}
\usepackage{xcolor}
\usepackage{multicol}
\usepackage{caption}
\usepackage{url}
\setcitestyle{aysep={}}

\usepackage{mathptmx}

\usepackage{geometry}
\usepackage[hyperfootnotes=false]{hyperref}
\hypersetup{
 colorlinks=true,
 linkcolor=blue!50!black,
 citecolor=blue!50!black,
 urlcolor=blue!50!black
 }


\date{}

\begin{document}

\title{A Node Embedding Framework for Model-Centric Conformance Checking in Process Mining}

\author{Mostafa Abbasi\orcidID{0000-0002-7479-6852} \and
Homayoun Najjaran\orcidID{0000-0002-3550-225X}}

\authorrunning{Abbasi and Najjaran}
\institute{Faculty of Engineering and Computer Science, University of Victoria, Victoria, BC, Canada\\
\email{\{abbasi,najjaran\}@uvic.ca}}

\maketitle

\begin{abstract}
Evaluating the quality of process models discovered from event logs is a fundamental challenge in process mining. Traditional evaluation relies on conformance metrics such as fitness and precision, which are highly sensitive to noise and deviations in event logs and therefore often fail to give a consistent assessment of discovery algorithms. We introduce a model-centric evaluation framework that combines node embeddings with Procrustes alignment to enable cross-graph comparability, since embeddings derived from different graphs reside in distinct vector spaces. The proposed PetriNetGraph method transforms Petri net process models into graph representations compatible with embedding algorithms. Comparing the node embeddings of ground-truth and discovered models through a parameterized cosine distance yields a unified and bounded metric for evaluating process discovery across datasets and methods. We evaluate the framework under static and context-aware scenarios to examine how prior knowledge of activity existence influences the assessment of structural similarity. The framework also integrates a compliance checking mechanism that localizes deviations at the activity level and an entropy-regularized node distance aggregation function that quantifies global deviations across the model. Experiments on nine publicly available event logs show that the approach provides a reliable embedding-based conformance indicator that reflects the behavior of fitness and precision. Experiments on synthetic logs with controlled noise levels confirm that it is substantially more robust to log deviations than traditional conformance metrics, PetriNet2Vec, and its dynamic-window and Procrustes-aligned variants, while remaining sensitive to process model degradation. All implementation details and datasets are publicly available.

\end{abstract}

\section{Introduction and Related Work}

A successful implementation of process mining relies on three essential pillars: i) high-quality event logs, ii) effective process model discovery methods, and iii) robust evaluation criteria for assessing model quality. These components are deeply interdependent and collectively determine the reliability and interpretability of process mining outcomes. Among them, event logs form the foundation for the extraction and refining process models, as they represent the observable behavior of business operations. However, in real-world settings, event logs are often noisy, incomplete, or inconsistent due to recording errors, system limitations, or operational inefficiencies \citep{math11132858}. As a result, achieving high log quality remains a significant challenge and frequently compromises the accuracy of both model discovery and evaluation. Traditional conformance checking techniques assess how closely an event log aligns with a given process model. Yet, their effectiveness is limited when logs deviate from idealized behavior, often leading to misleading conclusions about model quality.  

\subsection{Event log repair}
To address this limitation, Sommers et al. \cite{sommers2024assessing} introduced a ground-truth-based evaluation framework that uses synthetic process models and event logs enriched with controlled deviations. While valuable, such approaches still depend on expert-defined base models ($\mathcal{M}_0$) and manual noise injection, limiting scalability. This challenge motivates a novel evaluation perspective that remains reliable and scalable even when event logs contain deviations and noise.
To overcome these constraints, alternative techniques have emerged, leveraging generative and probabilistic methods to enhance both log quality and model robustness. Conditional variational autoencoders have also been employed to generate synthetic traces that preserve control-flow integrity and temporal consistency \citep{graziosi2024generating}. Furthermore, log repair methods based on Bayesian inference and stochastic Petri nets enable probabilistic reasoning over incomplete traces \citep{shahzadi2022repairing}. In addition, Hybrid Elusive Case Repair Method (HERE), a human-in-the-loop, Transformer-based approach that integrates rule-based logic and domain knowledge is another proposed approach to repair missing case identifiers in event logs, improving data quality and reliability in process mining \citep{zetzsche2025case}. Complementary data augmentation strategies have also been proposed to support reinforcement learning and improve process alignment in complex business environments \citep{abbasi2025forlapsinnovativedatadrivenreinforcement, Matteo2025}.

\subsection{Process model discovery advances}
A key frontier in process mining research involves improving the precision and robustness of process model discovery techniques. Notably, these advancements transcend traditional business contexts, contributing to user behavior analysis that informs user interface design across various systems~\citep{rehse2024user}. 

Both unsupervised and supervised learning techniques have been explored extensively to construct a process model $\mathcal{M}$ from an event log $\mathcal{L}$ \citep{sommers2023supervised}. Recent approaches in this domain primarily aim to strike a balance between achieving high precision and maintaining acceptable levels of fitness. However, simultaneously optimizing both objectives remains a challenging task~\citep{rennert2023improving}. With the rise of AI, recent work has introduced graph neural network (GNN)-based approaches for process discovery. These methods use structural learning to identify process patterns and optimize model construction. Evaluation typically relies on traditional conformance metrics such as fitness, precision, and F-score, while training objectives often aim to minimize the probability of selecting incorrect candidate places during the model construction process \citep{sommers2023supervised}. 

\subsection{Conformance metrics evaluation}
A third key area in process mining research focuses on evaluating the quality of models produced by state-of-the-art discovery methods. Four core conformance metrics are commonly assessed: fitness, precision, generalization, and simplicity~\cite{van2016data}. These metrics, often computed using token-replay or alignment-based techniques, provide a comprehensive evaluation of process model quality \cite{Carmona2022}. Ideally, for a process model $\mathcal{M}$ with behavior $\mathcal{B}(\mathcal{M})$ and an event log $\mathcal{L}$ with behavior $\mathcal{B}(\mathcal{L})$, we aim for behavioral inclusion: $\mathcal{B}(\mathcal{M}) \subseteq \mathcal{B}(\mathcal{L})$. Precision, which contrasts with fitness in interpretation, quantifies the degree to which a process model allows for only behaviors that were actually observed in the event log and it is typically calculated as $|\mathcal{B}(\mathcal{L}) \cap \mathcal{B}(\mathcal{M})| / |\mathcal{B}(\mathcal{M})|$ \citep{van2022process}. However, balancing these metrics is challenging, as improving one often comes at the expense of another. Moreover, both fitness and precision are highly dependent on event log quality. In practice, logs frequently contain noise, deviations, and incomplete traces, which can distort the evaluation of process models and undermine conformance assessments. This issue is compounded by log bias, which arises from data imperfections, and differs from representational bias, which stems from the modeling assumptions inherent in discovery algorithms \citep{van2011representational, math11132858}. Addressing both forms of bias is crucial to ensure fair, accurate, and robust evaluation of process discovery techniques. Several efforts have been made to improve the criteria used to evaluate process model quality. Fitness, for example, can be defined at different levels: at the case level, it refers to the fraction of traces in the log that can be fully replayed by the model; at the event level, it represents the fraction of events that are valid according to the model’s behavior \citep{van2011representational}. Other studies have been conducted to enhance various quality criteria, including fitness and precision. Karunaratne et al. \cite{karunaratne2024role} proposed a simplified approximation method for more effectively measuring generalization, addressing limitations in how event logs reflect underlying process behavior. Their work further explores the use of bootstrap-based resampling to estimate unseen behavior and enhance generalization assessment, offering a more nuanced view of model performance beyond traditional conformance metrics.

\subsection{Business process similarity evaluation}
Both fitness and precision refer to the similarity between an event log and a process model, albeit from different perspectives. Extending this notion, researchers have sought to quantify similarity between process models themselves. For instance, Tavares Garcia et al. \cite{GARCIA2023102211} introduced BPMN-Sim, a structural similarity technique for BPMN 2.0.2. Their method measures the similarity of subprocesses within larger processes by leveraging weighted subprocess hierarchies and dynamic vectorial signatures, producing normalized similarity scores [0~1] validated against practitioner evaluations. While effective for identifying similar subprocesses, this approach is not designed to compare complete process models against all traces of an event log. Its objective remains limited to detecting structurally similar subprocess fragments, rather than evaluating overall model–log correspondence.

Similarly, Salaün \cite{10043285} proposed two complementary techniques for measuring BPMN process model similarity: a syntactic approach based on local and global structural criteria, and a semantic approach based on comparing Labelled Transition Systems (LTS). They further developed an algorithm to relate semantic LTS models back to BPMN representations, enabling consistency across syntactic and semantic perspectives. However, their method, similar to \citep{GARCIA2023102211}, is restricted to comparing process models directly (whether structurally or by activity labels) without addressing the alignment between discovered models and the multiple process traces contained in an event log.

In a broader review, numerous functions have been proposed to quantify similarity between process models heuristically~\citep{schoknecht2017similarity}, highlighting the diversity of structural, syntactic, and semantic measures available. Yet, these contributions share a common limitation: they are not designed to establish trace-level similarity between event logs and discovered process models, which is a fundamental requirement for conformance checking. Consequently, while valuable for model-model comparison, these works are not directly applicable to evaluating structural similarity in the context of process mining.

\subsection{Embedding-based approaches}

Despite significant advances in generative AI and graph neural networks (GNNs), their application to process mining, particularly for process model evaluation, remains limited. Among these GNN applications, notable efforts include enhancing the explainability of prediction results using Gated Graph Neural Networks \citep{harl2020explainable}, applying graph transformers for remaining time prediction \citep{amiri2024pgtnet}, next activity prediction \citep{nguyen2025next, pfeiffer2025learning}. However, these works are primarily situated within the domain of predictive business process monitoring. 

In addition, research in related domains such as network analysis has extensively investigated methods for defining graph-, node-, and edge-level similarity. When adapted to the context of process models, however, the focus shifts toward assessing the similarity of activity behavior relative to the event log (\textit{i.e.}, the ground truth). This perspective is particularly critical, as it directly supports compliance checking by systematically identifying deviations between the discovered process model and the recorded behavior. In this regard, node embedding approaches \citep{saumya2025efficient}, and in particular DeepWalk \citep{perozzi2014deepwalk} and Node2Vec \citep{grover2016Node2Vecscalablefeaturelearning}, offer promising directions for evaluating the similarity between process models and event logs. These frameworks learn continuous node embeddings that preserve the local structural context ($\mathcal{N}_S(u)$). In addition, given the graph-like structure of process models, particularly Petri nets encoded in PNML (Petri Net Markup Language), such embeddings provide a mechanism for capturing meaningful activity representations.
While node embeddings have been successfully applied in diverse domains such as temporal knowledge graphs \citep{jia2025reasoning, sun2025rkr}, community detection \citep{cavallari2017learning}, node classification \citep{shen2020network}, and link prediction \citep{mallick2019topo2vec}, existing frameworks are typically tailored to learning representations from a single, large-scale graph.

Building on this idea, De Koninck et al. \cite{Koninck2018} applied the Continuous Bag-of-Words (CBOW) model to embed various process entities, such as activities, traces, event logs, and models, into a vector space. Similarly, in a more comprehensive study, Colonna et al. \cite{COLONNA2024200423} introduced PetriNet2Vec, a methodology for embedding entire Petri nets into vector representations to facilitate cluster analysis, as well as for learning embeddings of individual tasks. Notably, although they share the same embedding dimensionality, they are learned in different vector spaces. Consequently, the resulting scores cannot be directly interpreted as strict indicators of model-to-model or node-to-node similarity. In other words, while PetriNet2Vec effectively converts Petri nets into graph structures for embedding, the embeddings from separate models cannot be directly aligned for comparison across corresponding nodes. Nevertheless, this line of work highlights the potential of embedding-based Petri net representations for conformance checking, provided that meaningful cross-graph structural comparison can be achieved for robust model-centric evaluation. Beyond process-mining-specific embedding approaches, the broader graph similarity literature includes methods such as graph edit distance \cite{gao2010survey}, Weisfeiler and Leman graph kernels \cite{kriege2022weisfeiler}, and spectral geometry measures \cite{elghawalby2008measuring}. These methods are relevant as graph-level comparators, but they do not directly provide the node-level alignment and activity-level discrepancy analysis targeted by the present framework. For this reason, our empirical comparison focuses on the closest embedding-based Petri net baseline, while broader graph-level baselines remain an important direction for future evaluation.

\subsection{Research gaps and contributions}

To enable such comparison, node embeddings must be mapped into a shared embedding space; otherwise, direct comparison is infeasible. A further limitation of prior work lies in its reliance on pairs of consecutive tasks, which neglects long-term dependencies within the process model. This shortcoming stems from overlooking the crucial role of random walk generation in node embedding approaches. Random walks are indispensable for capturing higher-order dependencies among tasks, and their omission amplifies the effect of noise in event logs by disregarding essential structural context.

To address these shortcomings, we propose a node embedding-based framework that provides a reliable metric for process model evaluation. First, the metric is designed to reflect both fitness and precision simultaneously, effectively unifying them into a single measure. Second, it explicitly captures long-range dependencies between activities, accounting for the flexibility and variability that characterize real-world processes, where noises in the event log can reduce the accuracy of discovered models. To this end, we implement a tailored random walk mechanism for business processes, in which the walk length is adaptively determined by the structural properties of each process. 
In addition, since embeddings trained on separate models lie in different vector spaces, we further introduce an alignment step that translates embeddings into a shared space, enabling accurate cross-graph comparison. For measuring similarity, we adopt a modified cosine distance function, a widely used technique for comparing embeddings. This function operates across all nodes (both matched and unmatched), yielding a unified similarity score within a bounded range. To balance the contribution of individual nodes, we apply an adjustable and optimized weighting mechanism that regularizes the average distance across nodes. This flexibility ensures that evolving business processes are not constrained by a single fixed similarity threshold, but instead allows closer structural matches to be emphasized where appropriate.

Furthermore, the context of a process carries valuable information for the process analysis, especially in the application of AI to business process management \citep{gwinner2025context, rinderle2023predictive}. Hence, we evaluate the framework under two experimental scenarios. In the first, a context-aware setting, tasks not present in each event log trace (the ground truth) are filtered out, allowing more faithful alignment between log and model. In the second, a static setting, the same process model is used across all comparisons to test robustness.

Finally, to assess the sensitivity and robustness of our framework, we designed an additional experiment in which noise is injected into both event logs and process models. The goal is to ensure that the proposed metric is sensitive to deviations in the process model, yet robust to noise in the event log, which often reflects unavoidable real-world variability rather than systematic modeling errors.

Accordingly, the proposed study is situated between classical conformance evaluation and recent embedding-based approaches, using the former as a comparative baseline and the latter as the methodological foundation of the framework.

The main contributions of this paper are summarized as follows:

\begin{itemize}

\item Design and implement a generic PetriNetGraph constructor, enabling any discovered Petri net model to be transformed into a graph representation compatible with Graph Neural Network (GNN) architectures. This component bridges traditional process mining with modern AI-driven graph techniques, providing a flexible and scalable foundation for advanced model evaluation.

\item Introduce a customized adaptive random walk strategy that allows the framework to capture long-range dependencies between activities, reducing the impact of event log noise and strengthening the reliability of activity-to-activity relations.

\item Support compliance checking using node embedding at the activity level, enabling the detection of outliers and identification of misbehaving activities. To aggregate local node distances into a unified evaluation metric, we employ an entropy-regularized optimization mechanism. This ensures balanced weighting across nodes, while a tunable parameter allows the framework to be generalized for diverse process model evaluation scenarios.

\item Extend node embedding techniques beyond their conventional single-graph setting by introducing a cross-graph embedding distance function integrated with the skip-gram algorithm. This enables direct measurement of similarity between corresponding nodes in the ground truth event log and discovered process models. The proposed distance function provides a standardized, bounded similarity score that jointly reflects structural and behavioral alignment across models.

\end{itemize}

This paper is organized as follows. In Section \ref{method}, we present our proposed method and describe the preliminaries in detail. Section \ref{exp} provides detailed information about the datasets and the configuration settings of the models used in our experiments. The results and performance evaluation of our approach, in comparison with alignment-based conformance checking methods and benchmarks are discussed in Section \ref{results}. Finally, we conclude the paper and outline directions for future research in Section \ref{conclusion}.

\section{Preliminaries and Methodology}
\label{method}

Upon applying a process discovery algorithm to an event log $\mathcal{L}$, a corresponding Petri net $\mathcal{M}$ is derived, encapsulating the underlying process behavior such that $\mathcal{B}(M) \subseteq \mathcal{B}(L)$. The process model $\mathcal{M}$ is represented as a directed graph $G$, herein referred to as a \textit{PetriNetGraph}. 
Petri nets are directed bipartite graphs consisting of places \(P\), transitions \(T\), and arcs \(F\), where transitions may carry activity labels or represent silent (unobservable) steps \cite{van2022process}. In our framework, this standard Petri net structure is transformed into a derived graph representation for embedding and structural comparison. To support this computational representation, we introduce explicit source and sink marker nodes and distinguish labeled (i.e., $\mathcal{A}_M$, which may differ from the activity set in the event log $\mathcal{A}_L$) and silent transitions as separate node roles. These additions are not part of the formal definition of Petri nets, but are used only in the derived graph representation employed by the proposed method.

All possible sequences over $\mathcal{A}_M$, denoted as $\sigma^* \subseteq \mathcal{B}(M)$, are referred to as \textit{model traces}. Conversely, traces $\sigma \subseteq \mathcal{A}_L$, observed directly from the event log, define the log behavior $\mathcal{B}(L)$ and are referred to as \textit{truth traces} (\textit{i.e.,} ground truth). Comparing $\sigma^*$ with $\sigma$ enables evaluation of the fidelity of the model behavior $\mathcal{B}(M)$ relative to the actual behavior $\mathcal{B}(L)$.

Both approaches require transformation of $\mathcal{B}(L)$ and $\mathcal{B}(M)$ into graph-based structures. In this study, we generate embeddings for these graphs, which enables the comparison of discovered process models independent of the discovery method used. This embedding approach captures the structural and contextual properties of traces, supporting robust, trace-level comparison with the ground truth. Further preliminary concepts and representations underlying the proposed framework are introduced in Subsections~\ref{pgnet}-\ref{skipgram}, whereas Subsections~\ref{disfunc}-\ref{entraggr} present the methodological contributions of this work.

\subsection{PetriNetGraph construction}
\label{pgnet}
\begin{figure*}[tb]
\begin{minipage}[t]{0.47\textwidth}
\vspace{0pt}
\begin{algorithm}[H]
\caption{PetriNetGraph Construction}
\label{alg:PetriNetGraph}
\begin{algorithmic}
\Procedure{Initialize}{$\mathcal{L}, \mathcal{M}, M_0, M_f$}
  \State Initialize $G$ (PetriNetGraph)
  \State Call \textsc{ConstructPetriNetGraph}($\mathcal{M}$)
\EndProcedure
\Procedure{ConstructPetriNetGraph}{$\mathcal{M}$}
  \State $\mathcal{M} = (P, T, F), T = \mathcal{A}_M \cup \{\tau\}$
  \State $V \leftarrow \emptyset, E \leftarrow \emptyset$
  \ForAll{$p \in P$}
    \State $V \leftarrow V \cup \{p\}$
  \EndFor
  \ForAll{$a_M \in T$}
    \State $V \leftarrow V \cup \{a_M\}$
  \EndFor
  \ForAll{$(x, y) \in F$}
    \State $E \leftarrow E \cup \{(x, y)\}$
  \EndFor
  \State \Return $G = (V, E)$
\EndProcedure
\Procedure{ContextAware}{$\sigma, G$}
  \State Clone $G$ as $G_\sigma$
  \ForAll{\( a_M \in V_\sigma \) such that \( a_M \notin \sigma \)}
    \State $V_\sigma \leftarrow V_\sigma \setminus \{ a_M \},$  \quad
    \State $E_\sigma \leftarrow E_\sigma \setminus \{ (u, a_M), (a_M, w) \in E_\sigma \}$
    \If{$\exists v_\sigma \in V_\sigma $ such that \( \deg(v_\sigma) < 2 \)}
      \State $V_\sigma \leftarrow V_\sigma \setminus \{v_\sigma\},$ \quad
      \State $E_\sigma \leftarrow E_\sigma \setminus \{ (u,v) \in E_\sigma \text{ or } (v,w) \in E_\sigma \}$
      \State Repeat
    \EndIf
  \EndFor
  \State \Return $G_{\sigma^*}$
\EndProcedure
\Procedure{GenerateTraceGraphs}{$\mathcal{L}$}
  \State Initialize empty list $T_G$
  \ForAll{traces $\sigma$ in $\mathcal{L}$}
    \State $G_\sigma \gets$ \textsc{ContextAware}($\sigma, G$)
    \State Append $G_\sigma$ to $T_G$
  \EndFor
  \State \Return $T_G$
\EndProcedure
\end{algorithmic}
\end{algorithm}
\end{minipage}%
\hfill
\centering
\hspace{0.02\textwidth}
\begin{minipage}[t]{0.48\textwidth}
\vspace{0pt}
\captionsetup{type=figure}
\begin{subfigure}[t]{\linewidth}

\includegraphics[width=\linewidth]{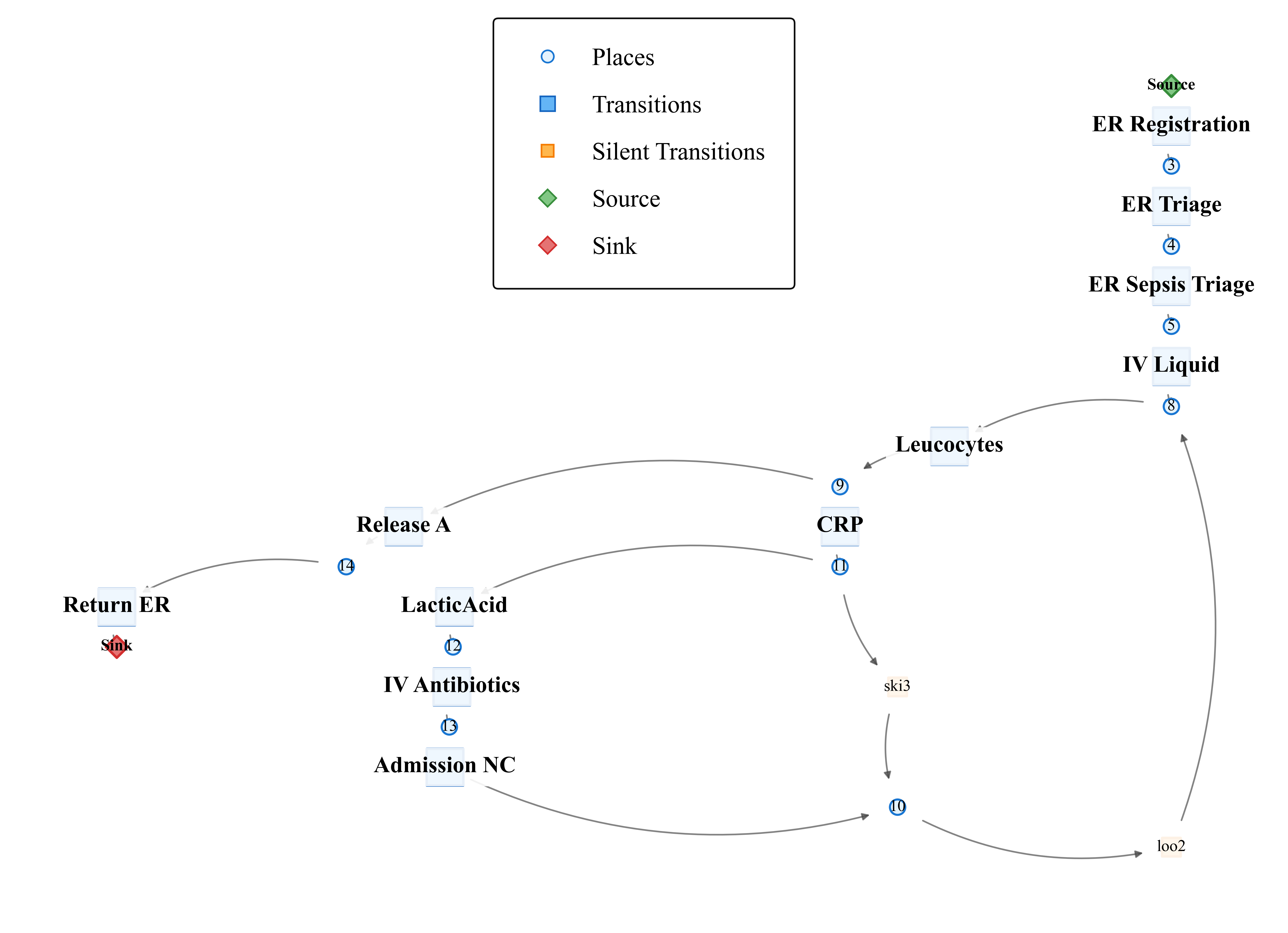}
\caption{Ground truth trace}
\label{fig:GTSEPSIS}
\end{subfigure}

\begin{subfigure}[t]{\linewidth}
\includegraphics[width=\linewidth]{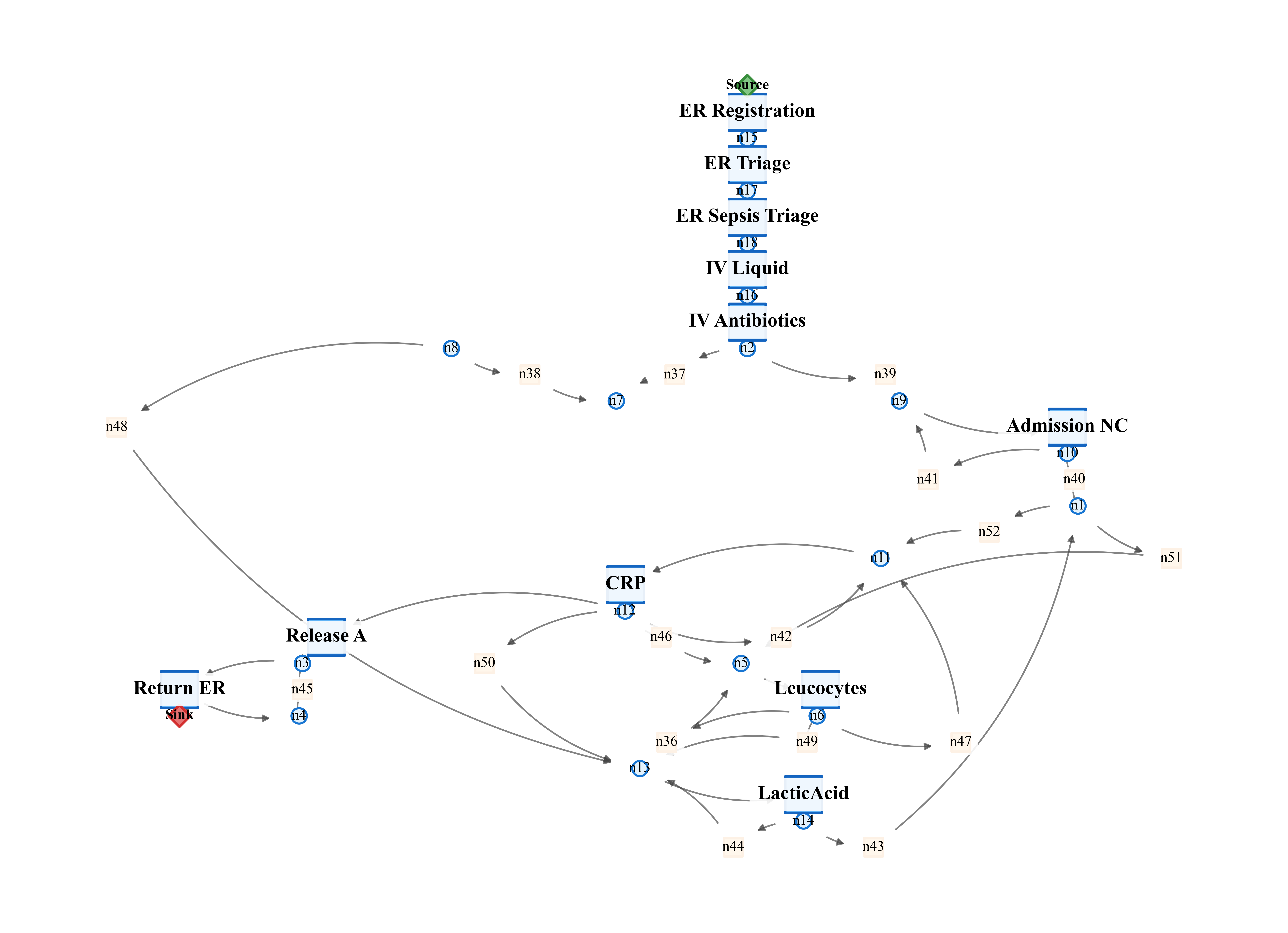}
\caption{Process model trace}
\label{fig:SepsisTraces}
\end{subfigure}
\caption{Constructed graph representation of a Petri net for a Sepsis case.}
\label{fig:SepsisTracesall}

\end{minipage}
\end{figure*}

To convert event logs and Petri net models in PNML format into directed graphs suitable for embedding, we define a structured procedure for parsing PNML files, constructing graph representations, and preparing trace-level data. Algorithm~\ref{alg:PetriNetGraph} presents the initialization and transformation pipeline. First, the Petri net model $\mathcal{M}$ is translated into a graph $G=(V,E)$ containing places, transitions, and their flow relations. Next, for each trace $\sigma \in \mathcal{L}$, a context-aware refinement is applied: transitions not present in $\sigma$ are removed along with structurally disconnected nodes, yielding a trace-specific subgraph $G_{\sigma^*}$. This produces a collection of trace graphs $T_G$ that preserve contextual execution information.

We adopt two complementary strategies for this transformation. The first, the Static Model Approach, treats $\mathcal{B}(M)$ as a fixed behavioral representation where all model-supported traces are valid ($\sigma^* \subseteq \mathcal{B}(M)$). The second, the Context-Aware Filtering Approach, adapts the model per trace by retaining only transitions relevant to the observed execution ($\sigma^* \subseteq |\sigma \cap \mathcal{B}(M)|$). This context-sensitive refinement ensures alignment of the model with actual trace behavior, thereby improving the accuracy of similarity comparisons and embedding generation.

Our experimental resources and all the implementations are publicly available.\footnote{\url{https://github.com/mstabc/cc_node2vec}}

Fig. \ref{fig:SepsisTracesall} illustrates the ground truth and process model trace graph representations for a specific case ID in the Sepsis dataset.

\subsection{Random walk mechanism}

To effectively apply node embedding via random walks to accommodate the unique structural and behavioral properties of Petri nets. This adaptation involves generating random walks that (1) preserve the directionality inherent in process flows, (2) respect the bipartite nature of the graph, and (3) capture complex behavioral constructs such as parallelism and loops that are important characteristics of process models.

To capture meaningful structural and sequential information, we employ a random walk strategy, as illustrated in Algorithm \ref{randomwalk}. The random walk is a stochastic process that traverses the graph by selecting a node and iteratively moving to one of its successors. Given a directed graph $G$, the random walk process is initiated from a chosen node and progresses for a specified walk length ($l$) depending on the maximum process trace ($\sigma \in (\mathcal{L} \cup \mathcal{M})$. The transition to the next node is determined by randomly selecting from the current node’s successors. If a node lacks outgoing edges, the walk terminates prematurely.

This approach generates all possible traces by incorporating places ($p$) and silent transitions ($\tau$), enabling comprehensive exploration of the process model through graph walks. However, to construct meaningful walks, the selected transitions should structurally align with the ground truth trace while leveraging the flexibility provided by silent transitions and places to traverse the graph. For a given dataset consisting of multiple graphs ($T_G$), the generate walks function (Algorithm \ref{alg:generate_walks}) executes $r$ walks per node, ensuring sufficient coverage of node interactions. To focus on meaningful structural elements, only nodes belonging to predefined categories (sink, source, and transition) are selected as starting points. And it can be applied for both $\sigma$ and $\sigma^*$.

Unlike in undirected graphs, where random walks can move freely or return arbitrarily, our approach constrains transition choices based on the execution logic of the process, thereby preserving model compliance. This adjustment ensures that random walk generation reflects realistic operational flows and maintains the semantics inherent to business processes.

Once node traversal sequences are generated, we apply the Word2Vec algorithm to embed the nodes into a continuous vector space. Specifically, we use the skip-gram model ($sg=1$), which is well-suited for learning representations in which the context, defined by neighboring nodes in the walk, encodes meaningful semantic relationships.

\begin{figure*}[t]
\begin{minipage}[t]{0.48\textwidth}
\begin{algorithm}[H]
\small
\caption{Directed Random Walk on a Process Graph}
\label{randomwalk}
\begin{algorithmic}
\Procedure{RandomWalk}{$G, u, l$}
  \State Initialize $walk \gets$ Empty List
  \State $current\_node \gets u $
  \For{$i = 1$ to $l$}
    \State Append $(current\_node)$ to $walk$
    \State $neighbors\gets$successors of $current\_node$
    \If{$neighbors = \emptyset$}
      \State \textbf{break}
    \EndIf
    \State $current\_node \gets$ sample from $neighbors$
  \EndFor
  \State \Return $walk$
\EndProcedure
\end{algorithmic}
\end{algorithm}
\end{minipage}%
\hfill
\begin{minipage}[t]{0.48\textwidth}
\begin{algorithm}[H]
\small

\caption{Generate Directed Random Walks}
\label{alg:generate_walks}
\begin{algorithmic}
\Procedure{GenerateWalks}{$\mathcal{G}, r, l$}
  \State Initialize $walks \gets$ Empty List
  \ForAll{$(case\_name, G) \in \mathcal{G}$}
    \ForAll{nodes $u \in |\mathcal{A}_M\cup M_0 \cup M_f| \subset G$}
      \For{$i = 1$ to $r$}
        \State $walk \gets$ \Call{RandomWalk}{$G, u, l$}
        \State Append $walk$ to $walks$
      \EndFor
    \EndFor
  \EndFor
  \State \Return $walks$
\EndProcedure
\end{algorithmic}
\end{algorithm}
\end{minipage}
\end{figure*}

\subsection{Skip-gram}
\label{skipgram}
After deriving Petri net representations from event logs and corresponding process models, we apply a node embedding algorithm to learn continuous feature representations for the nodes in these models. Node2Vec or DeepWalk are scalable graph embedding frameworks that capture both local and global structural characteristics via biased and unbiased random walks, making them well-suited for encoding complex process behaviors.

Each generated random walk is treated as a sequence, analogous to a sentence in natural language processing. We then employ the skip-gram model from Word2Vec to learn node embeddings, where the objective is to maximize the likelihood of predicting a node’s neighborhood within a predefined context window. This formulation effectively captures the functional and temporal context in which each transition or activity occurs within the process.

Using the structural expressiveness of Petri nets and adapting skip-gram to reflect process semantics, our approach produces embeddings that encapsulate the dynamic behavior of business processes. These learned representations can support a variety of downstream tasks, such as process variant classification, anomaly detection, and clustering.
In this method, the function ($f$) that gives a vector for every node learns node embeddings so that nodes can predict their random-walk neighbors ($\mathcal{N}_A(u)$) \citep{grover2016Node2Vecscalablefeaturelearning}.

\begin{equation}
    \label{lognode}
    \max_f \sum_{u \in V} \log \Pr(\mathcal{N}_A(u) \mid f(u))
\end{equation}

This will consider the whole graph, however, to avoid deviation and error in the $\mathcal{L}$, we can break it into multiple independent neighborhoods and combine them (Equation \ref{independantnodes}).

\begin{equation}
\label{independantnodes}
\Pr(\mathcal{N}_A(u) \mid f(u)) = \prod_{n_i \in \mathcal{N}_A(u)} \Pr(n_i \mid f(u))
\end{equation}

The probability of $n_i$ (\textit{i.e.,} transitions being a neighbor of $u$ (indirect neighbor by places)) depends on the dot product between their embeddings as indicated in Equation \ref{softwaxfn}. Where \( f(n_i) \cdot f(u) \) is the dot product of the embeddings, indicating their similarity.
\begin{equation}
\label{softwaxfn}
    \Pr(n_i \mid f(u)) = \frac{\exp(f(n_i) \cdot f(u))}{\sum_{v \in V} \exp(f(v) \cdot f(u))}
\end{equation}
This means that activities that co-occur in $\sigma$ and $\sigma^*$ will have higher similarity scores, reinforcing that embedding proximity can reflect behavioral alignment, thus serving as a data-driven indicator of conformance.

\subsection{Distance function}
\label{disfunc}

In this work, we employ a widely used metric, cosine distance, to measure distance or similarity between corresponding nodes across two process model graphs. In process model discovery, embedding similarity between the embeddings of corresponding labels in $\mathcal{A_L} $ and $ \mathcal{A_M}$ offers a meaningful proxy for conformance. When both exhibit similar neighborhood contexts, their embeddings become closely aligned. This reflects high fitness (capturing observed behavior) and precision (avoiding unobserved behavior).

Cosine distance measures the angular difference between two vectors, focusing purely on their orientation rather than magnitude. This makes it particularly suitable in high-dimensional spaces where the direction of vectors carries more semantic meaning than their scale. Given two vectors \( \mathbf{u}, \mathbf{v} \in \mathbb{R}^d \), the cosine distance is defined as:

\begin{equation}
d_{\text{cos}}(\mathbf{u}, \mathbf{v}) = 1 - \frac{\sum_{i=0}^{n-1} u_i v_i}{\sqrt{\sum_{i=0}^{n-1} u_i^2} \cdot \sqrt{\sum_{i=0}^{n-1} v_i^2}}
\end{equation}

This distance ranges from 0 (identical direction) to 2 (opposite direction). Because cosine distance is scale-invariant, it is widely used for comparing node embeddings derived from random walk–based methods (\textit{e.g.,} Node2Vec, Deepwalk).

However, embeddings trained independently on two different but structurally similar graphs (even if the graphs are isomorphic) will lie in different vector spaces due to differences in initialization, optimization, and random walk paths. Consequently, even if the graphs are equivalent, the vectors will not be directly comparable.

\subsection{Procrustes alignment}

To resolve the incomparability of nodes in different embedding spaces, one possible strategy in the literature is to use a \textit{doc2vec}-style approach \citep{COLONNA2024200423}, where process model traces and ground-truth traces are jointly embedded into a single space. In such methods, each graph is treated as a document, and nodes are embedded with reference to both their local context and their global graph identifier, effectively producing a static vector representation of the process model. While attractive in principle, this approach is not suitable in our setting for several reasons. Since it produces a \textit{single} embedding per node (Activities (\textit{i.e.,} $\mathcal{A_L} $ and $ \mathcal{A_M}$)), it eliminates the fine-grained node-level representation necessary to capture structural and behavioral differences across models.

To resolve these shortcomings, a post-hoc alignment is needed to bring the embeddings into a shared space. One effective approach is to learn a \textbf{linear transformation} (\textit{i.e.,} \textbf{Procrustes alignment}) using a set of known node correspondences. 
Let \( \mathbf{E}_A \in \mathbb{R}^{n \times d} \) and \( \mathbf{E}_B \in \mathbb{R}^{n \times d} \) be matrices of node embeddings from Graph A and Graph B, respectively, for the same and common \( n \) nodes.

We then find an orthogonal transformation \( \mathbf{W} \in \mathbb{R}^{d \times d} \) that minimizes the Frobenius norm:

\begin{equation}
\min_{\mathbf{W}} \left\| \mathbf{E}_A \mathbf{W} - \mathbf{E}_B \right\|_F
\end{equation}

This optimization has a closed-form solution and yields a rotation matrix \( \mathbf{W} \) that best aligns \( \mathbf{E}_A \) to \( \mathbf{E}_B \). Once computed, the matrix \( \mathbf{W} \) can be applied to all embeddings in Graph A:

\begin{equation}
\mathbf{E}_A^{\text{aligned}} = \mathbf{E}_A \mathbf{W}
\end{equation}

After alignment, the node embeddings from Graph A and Graph B lie in the same latent space, allowing for meaningful distance or similarity computations, such as cosine distance. The quality of this alignment depends on the set of shared anchor nodes \( \mathcal{S} = \mathcal{A}_L \cap \mathcal{A}_M \) used to estimate \( \mathbf{W} \). These anchors determine the alignment primarily within the subspace they span, so the resulting transformation is most directly supported on the matched anchor nodes themselves. In practice, this provides a well-defined basis for comparing the shared structure across the two models, while the treatment of unmatched nodes is handled separately through the penalty mechanism.

\subsection{Entropy-regularized node distance aggregation}
\label{entraggr}
In practical process mining scenarios, graph comparison frequently involves models with non-identical node sets. This discrepancy arises because discovery algorithms often apply simplification or frequency-based filtering, which can remove infrequent or noisy activities, thereby reducing the set of represented nodes. As a consequence, the set of activity labels in the discovered model \( \mathcal{A_M} \) is generally a subset of the label set in the ground truth model \( \mathcal{A_L} \), such that \( \mathcal{A_M} \subseteq \mathcal{A_L} \). On the other hand, if the number of generated random walks is insufficient to traverse all relevant paths in the truth traces, certain nodes may remain uncovered. As a result, embeddings derived from such walks may fail to capture corresponding nodes across different graphs.

More generally, when comparing any two graphs with differing node sets, a challenge arises: how should we quantify structural dissimilarity when some nodes in the expected set lack a corresponding embedding? To address this, we introduce a penalty mechanism that reflects the absence of expected nodes. Specifically, when a node label \( \ell \in \mathcal{A_L} \) is missing in the trace model \( \mathcal{A_M} \), we impose a penalty proportional to the maximum observed node discrepancy across the compared sets. This penalty is further scaled by a factor 
$\alpha \in [0,1]$ which serves as a tunable parameter to control the impact of missing nodes on the overall similarity score. 
The penalty assigned to unmatched nodes is a modeling choice intended to capture the absence of expected structure in a simple and comparison-dependent manner. In the proposed formulation, an unmatched node receives the penalty \( \alpha \cdot M \), where \(M\) is the maximum observed distance among the matched nodes in the same comparison. This avoids using a fixed constant penalty that would be insensitive to the scale of the current model-log discrepancy, while preserving a direct and interpretable connection to the observed mismatch structure. The parameter \(\alpha\) controls the strength of this penalty explicitly. This default penalty should be interpreted as a general structural treatment of missing nodes, rather than as an attempt to infer their business criticality automatically. In public benchmark datasets, activities are not annotated as critical or non-critical, and a learned penalty would require external supervision that is not consistently available. When such domain knowledge exists, however, the framework can incorporate activity-specific importance through the node-level weighting terms, thereby allowing decision-makers to place greater emphasis on omissions involving critical activities.

Let:

\begin{multicols}{2}
\begin{itemize}\footnotesize
  \item $\mathcal{A}$: the set of expected labels (nodes)
  \item $\mathcal{A_M} \subseteq \mathcal{A}$: labels present in the trace model
  \item $\mathcal{A_L} \subseteq \mathcal{A}$: labels present in the ground truth model
  \item $\mathcal{S} = \mathcal{A_L} \cap \mathcal{A_M}$: the set of shared labels
  \item $\mathbf{v}_\ell^{(A_M)}$: embedding of label $\ell$ in the trace model
  \item $\mathbf{v}_\ell^{(A_L)}$: embedding of label $\ell$ in the ground truth model
  \item $W \in \mathbb{R}^{d \times d}$: orthogonal Procrustes alignment matrix
  \item $d(\cdot, \cdot)$: distance function (e.g., cosine)
  \item $\alpha$: penalty scaling factor
  \item $M$: maximum distance value for matched nodes
\end{itemize}
\end{multicols}

We define the penalized distance \( d_v \) for each node \( v \in \mathcal{A} \) as:
\begin{equation}
d_v =
\begin{cases}
d\!\left( W\,\mathbf{v}_v^{(A_M)},\, \mathbf{v}_v^{(A_L)} \right), & \text{if } v \in \mathcal{S},\\[4pt]
\alpha\, M, & \text{otherwise.}
\end{cases}
\end{equation}

We can compute a weighted aggregate using an entropy-regularized objective that minimizes a trade-off between the weighted sum of distances and the entropy of the weighting distribution:

\begin{equation}
\min_{\mathbf{z}} \sum_{v \in \mathcal{A}} z_v d_v + \lambda \sum_{v \in \mathcal{A}} z_v \log z_v
\end{equation}

subject to:

\begin{equation}
\sum_{v \in \mathcal{A}} z_v = 1, \quad z_v \geq 0 \quad \forall v \in \mathcal{A}
\end{equation}

This formulation enables a flexible trade-off between focusing on well-aligned core activities and considering the entire set of expected labels. Specifically:

\begin{itemize}
    \item When \( \lambda \to 0 \), the solution concentrates weight on the node(s) with the smallest distances, favoring minimal-case matching. This is appropriate when only partial alignment is necessary, such as in evolving or modular processes.
    \item When \( \lambda \) is large, the weight distribution becomes uniform, effectively recovering the standard average distance across all nodes, ensuring fairness in the evaluation.
\end{itemize}

As \( \lambda \to \infty \), the entropy regularization term \( \sum_{v \in \mathcal{A}} z_v \log z_v \) dominates the objective function, and the minimum is attained when the weights form a uniform distribution, i.e., \( z_v = \frac{1}{|\mathcal{A}|} \), which maximizes the entropy under the constraint \( \sum_{v \in \mathcal{A}} z_v = 1 \).

Given the optimized weights \( z_v \) over all nodes \( v \in \mathcal{A} \), the weighted structural distance between the trace model and the ground truth can be expressed as:

\begin{equation}
\mathbf{D}_{\text{weighted}} = \frac{1}{M} \sum_{v \in \mathcal{A}} z_v \cdot d_v
\end{equation}

Each node's contribution to the overall similarity can be modulated based on its relevance (via \( z_v \)), while still respecting the penalty scaling factor \( \alpha \) and the maximum distance bound \( M \).

The final distance metric \( D_{\text{weighted}} \) lies in the range \([0, 1]\), where lower values indicate greater structural similarity, while values approaching \(1\) indicate severe mismatches between the models or maximum dissimilarity. Importantly, \( D_{\text{weighted}} \) is not an arithmetic combination of fitness and precision, nor is it intended to replace their established individual interpretations. Instead, it serves as an embedding-based structural proxy for conformance that empirically reflects the behavioral information conveyed by both dimensions through a single distance measure. Specifically, fitness reflects the extent to which the discovered model can reproduce the behavior observed in the event log, whereas precision reflects the extent to which the discovered model avoids behavior not supported by the log. By comparing the structural contexts of corresponding nodes across the reference and discovered process models, \( D_{\text{weighted}} \) provides a unified view of their joint behavioral consistency without requiring these two measures to be computed separately. For reference, the F1-score is commonly used as a standard combined summary of fitness and precision, defined as their harmonic mean. However, the objective of the proposed metric is not to reproduce the F1-score directly, but rather to provide a structural conformance indicator whose behavior is empirically consistent with both traditional conformance dimensions and, by extension, with their combined trend.

\begin{figure}[htbp]
    \centering
    \includegraphics[width=\textwidth, height=0.9\textheight, keepaspectratio]{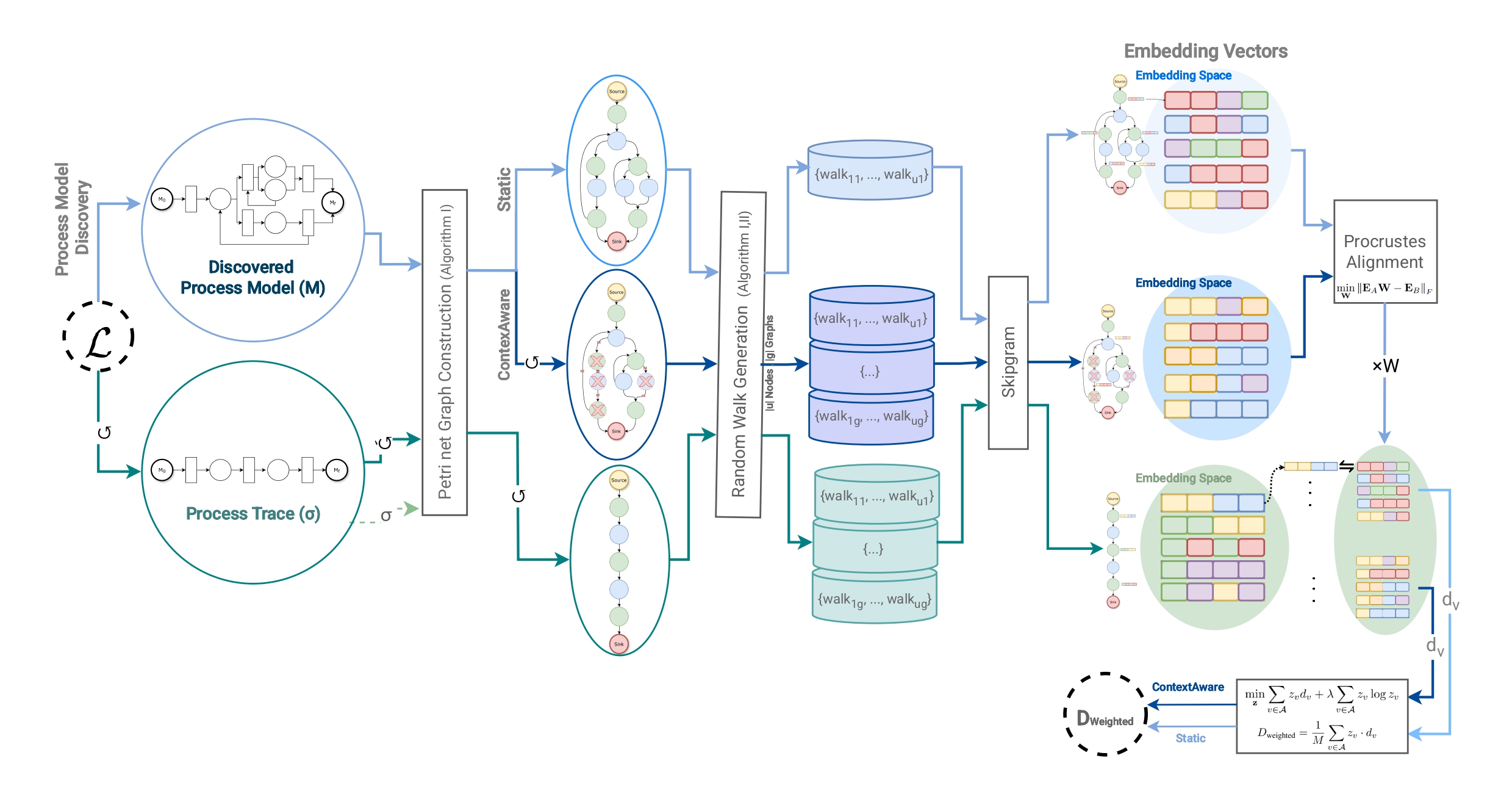}
    \caption{Detailed overview of embedding-based distance for process model discovery method evaluation}
    \label{framework}
\end{figure}

In Fig.~\ref{framework}, we outlined the overall procedure of the proposed approach. Given an event log $\mathcal{L}$, a process model $\mathcal{M}$ is discovered using any process model discovery method (\textit{e.g.}, ILP, GNN). For each case in $\mathcal{L}$, Petri net construction is applied to transform the PNML into a graph representation. Under the context-aware scenario, this procedure is repeated for all cases, where the set of activities in each trace ($\sigma$) is provided to $\mathcal{M}$ to filter the corresponding activities. Once the graph representations are obtained, random walks are generated and processed with the skip-gram algorithm (Section~\ref{skipgram}) to learn node embeddings. Procrustes alignment is then employed to project both context-aware and static embeddings into the ground-truth embedding space. Finally, entropy-regularized node distance aggregation is applied to compute the structural distances of the static and context-aware process models.

\section{Datasets and Experimental Setup}
\label{exp}

We utilized nine publicly available event logs together with their corresponding discovered process models, each represented as a Petri net in PNML format. These models were not generated within this study; instead, we adopted the benchmark models released by Sommers et al. \citep{sommers2023supervised} and evaluated them using the proposed framework. The selected datasets cover a variety of domains, including healthcare, construction, permitting, and financial processes, and comprise BPIC\_2012, BPIC\_2017, multiple variants of BPIC\_2020, Road Traffic Fines, and Sepsis. For comparison with traditional conformance measures, we explicitly report that fitness and precision were computed using alignment-based methods in the evaluation.

Our approach was implemented and evaluated across five distinct discovery methods: \textit{Inductive Miner~\cite{Leemansinductive}, ILP Miner \cite{van2008process}, Heuristics Miner \cite{weijters2011flexible}, Split Miner \cite{augusto2019split}, and GNN Miner \cite{sommers2023supervised}}.

Additional hyperparameters are configured to enhance the quality of the learned embeddings. The window size is dynamically set to half of the random walk length (corresponding to the maximum global length of the trace), defining the maximum distance between a node and its surrounding (preceding and succeeding) nodes within each walk. A smaller window captures localized structural dependencies, whereas a larger one reflects broader contextual relationships. By adopting a half-walk window, the model effectively captures global dependencies by encompassing both preceding and succeeding nodes along the walk. To improve computational efficiency and representation quality, negative sampling is applied with 1 negative node selected for each training instance, avoiding the need to compute full softmax probabilities. The dimensionality of the learned node embeddings is set to \( d = 128 \), which we use as a practical compromise between representational capacity and alignment stability. Very small dimensions may underrepresent structural context, whereas larger dimensions increase the ratio \( d / |\mathcal{S}| \) and therefore make the Procrustes alignment more weakly constrained when the shared anchor set is small. To assess the sensitivity of this choice, we conducted a parameter analysis over \( d \in \{12, 32, 64, 128\} \), together with \(\alpha\) and minimum count. The results, reported in Appendix \ref{paramanalysis}, show only modest variation in the residuals and no material change in the overall comparative conclusions.
In the configuration used in this study, the generated walk corpus is sufficiently rich that \texttt{min\_count} does not materially filter the embedding vocabulary. This is appropriate for an evaluation-focused framework, where node omission should be driven as much as possible by the discovery algorithm rather than by the embedding stage.

By integrating random walk-based sampling with Word2Vec embeddings, the method efficiently encodes the graph structure into a meaningful vector space. Within this framework, the skip-gram model aims to predict the surrounding context of a given node. Negative sampling enhances this learning process by introducing artificially generated negative example nodes that are not true neighbors and training the model to distinguish between actual context nodes and these random non-neighbors.

\subsection{Synthetic log}

To evaluate the robustness and performance of the embedding-based framework, we assess its ability to detect deviations from the ground truth by applying it to synthetic event logs. The objective is to compare the learned behavior from each discovered model with the original process, particularly under varying noise levels in both incremental and decremental scenarios. Synthetic traces are generated by probabilistically enforcing a set of predefined behavioral rules (e.g., precedence, co-occurrence, mutual exclusion), which define a structured prefix. This prefix is then extended with a variable number of randomly sampled, non-conflicting activities to complete the trace.

Using this setup, we generate the original event log $\mathcal{L}$ and apply a process discovery algorithm to obtain a corresponding process model $\mathcal{M}$. To simulate process anomalies, we introduce controlled deviations through a noise injection function, governed by a predefined set of noise levels $\varepsilon \in \{0\%, 10\%, \dots, 90\%\}$, resulting in a series of noisy logs $\mathcal{L}_\varepsilon$. Each noisy log $\mathcal{L}_\varepsilon$ exhibits behavior $\mathcal{B}(\mathcal{L}_\varepsilon)$, from which a model $\mathcal{M}_\varepsilon$ is discovered. This process enables the evaluation of node embedding in identifying behavioral divergence between the ground truth model and models derived from deviated logs.

\section{Results and Discussions}
\label{results}

In this section, we present an analysis of the evaluation results by comparing traditional conformance metrics, alignment-based fitness and precision, as reported in \citep{sommers2023supervised}, with the embedding-based distance measures.

This comparison aims to assess the effectiveness of the embedding method in capturing behavioral deviations in process models. As discussed earlier, we benchmark several widely-used process discovery algorithms, including Graph Neural Network Miner, Inductive Miner, ILP, Heuristics Miner, and Split Miner. By aligning structural embeddings with conventional metrics, we aim to investigate whether embedding-based distance can serve as a robust and complementary metric for evaluating process model quality across different levels of noise and discovery methods.

\subsection{Static model and context-aware model}
\label{stcoresult}
As detailed in Section \ref{method}, we evaluate the similarity between the discovered process model and the ground truth by comparing the trace sets $\sigma^*$ and $\sigma$. In the static model approach, all possible sequences over the activity set $\mathcal{A}_{M}$ are considered, i.e., $\sigma^* \subseteq \mathcal{B}_{M}$. Alternatively, in the context-aware approach, the discovered process model is refined based on the observed trace $\sigma$, leading to a more constrained comparison: $\sigma^* \subseteq |\sigma \cap \mathcal{B}_{M}|$. In this section, we report the results of both average embedding-based distances and traditional conformance checking measures (fitness and precision) to assess the behavioral alignment between discovered and ground truth models.

\begin{figure}[htbp]
    \centering
    \includegraphics[width=\textwidth, height=0.9\textheight, keepaspectratio]{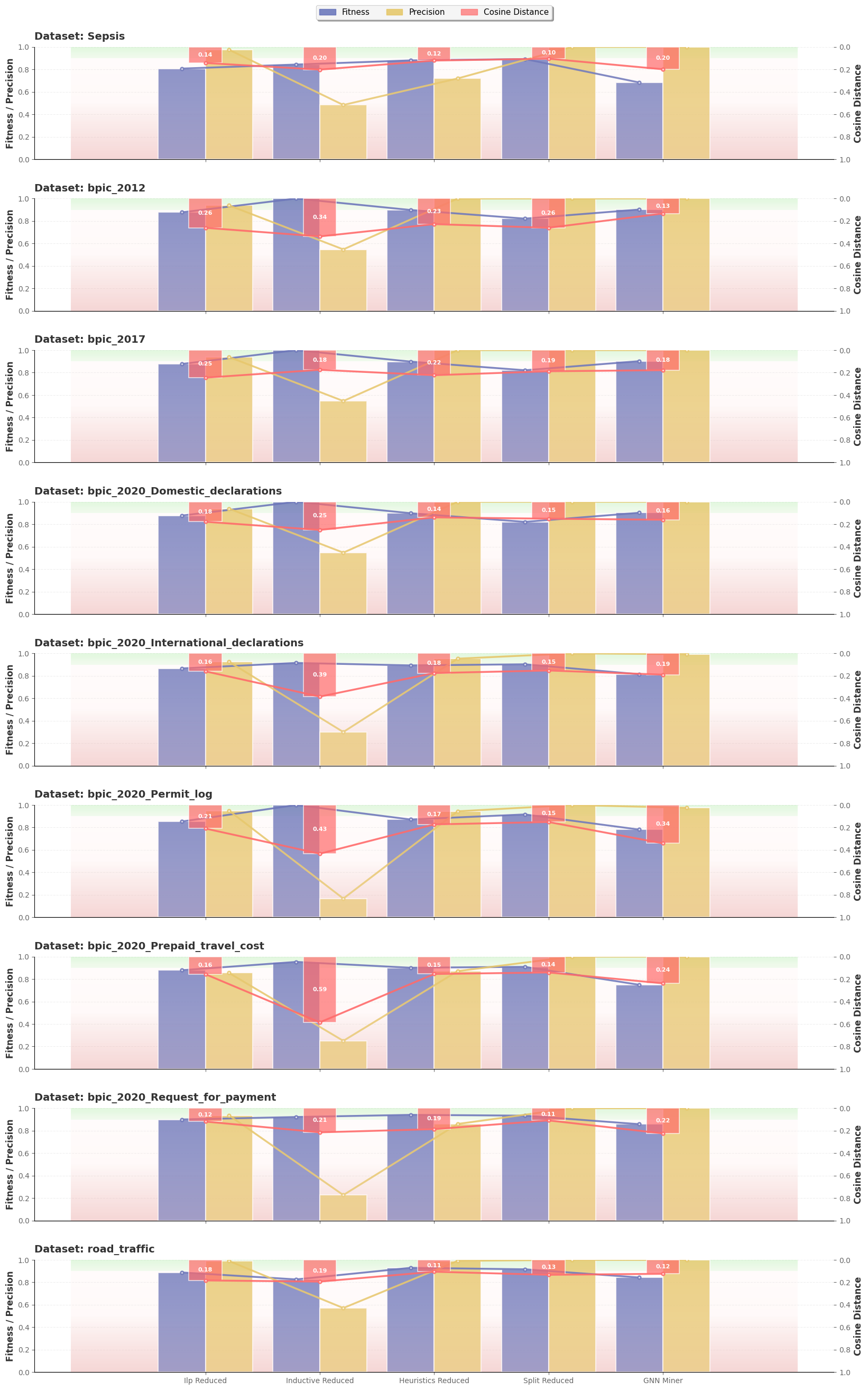}
    \caption{Static process model conformance checking results vs. cosine distance}
    \label{fig:minedcomp}
\end{figure}

Fig. \ref{fig:minedcomp} presents the cosine distance alongside fitness and precision scores obtained using the static model approach. In the main evaluation, we report results for the setting \(\lambda = \infty\) and \(\alpha = 1\), which corresponds to the most discrepancy-sensitive setting among the tested configurations. For completeness, Appendix~\ref{paramanalysis} reports results across multiple \(\lambda\) and \(\alpha\) values. As expected, settings with smaller \(\lambda\) reduce the influence of larger node-level discrepancies on the final aggregation, while smaller \(\alpha\) reduces the contribution of unmatched nodes. Although the absolute values of \(D_{\text{weighted}}\) change across these settings, the overall comparative patterns remain stable.

Consistent with fitness and precision, the accuracy of the results is influenced by both data availability and the performance of the process discovery algorithm. The left vertical axis in Fig. \ref{fig:minedcomp} represents fitness and precision (ranging from 0 to 1), where higher values indicate stronger conformance. Conversely, the right vertical axis shows cosine distance, plotted top to bottom, where lower values denote greater structural similarity. To aid interpretation, shaded regions are included: green corresponds to favorable results (fitness and precision exceeding 85\%), while red indicates less desirable outcomes. Endpoints within the green region signify satisfactory conformance performance. In addition, a line chart complements the bar plots by illustrating overall indicator trends, underscoring the alignment of metrics: higher fitness and precision correspond to lower cosine distance. For methods and datasets with acceptable fitness and precision (green region), the cosine distance remains within the 0.1–0.3 range, suggesting that the graph embeddings preserve semantic similarity even after the embedding process.

As expected, the cosine distance reflects the behavior of fitness and precision; when fitness and precision are lower, the distance increases, indicating greater structural divergence between the discovered model and the ground truth. Method performance varies across datasets. For example, in the Sepsis dataset, Split Miner achieves the lowest cosine distance (0.1), outperforming all other methods. This is followed by Heuristics Miner, ILP Miner, Inductive Miner, and GNN Miner. While assigning exact weights to balance fitness and precision in ranking is not feasible, cosine distance proves to be a reliable complementary metric, as it captures structural similarity and reflects both conformance dimensions.

Overall, Fig. \ref{fig:minedcomp} shows that methods such as Split Miner and Heuristic Miner, which already perform well in terms of fitness and precision, achieve cosine distances in the 0.1-0.2 range, thereby producing high-quality process models. By contrast, Inductive Miner demonstrates noticeably lower precision, and in cases where high fitness cannot compensate for this, cosine distances rise substantially (0.3-0.6), indicating considerable deviations from the ground truth.

\begin{figure}[htbp]
    \centering
    \includegraphics[width=0.7\linewidth]{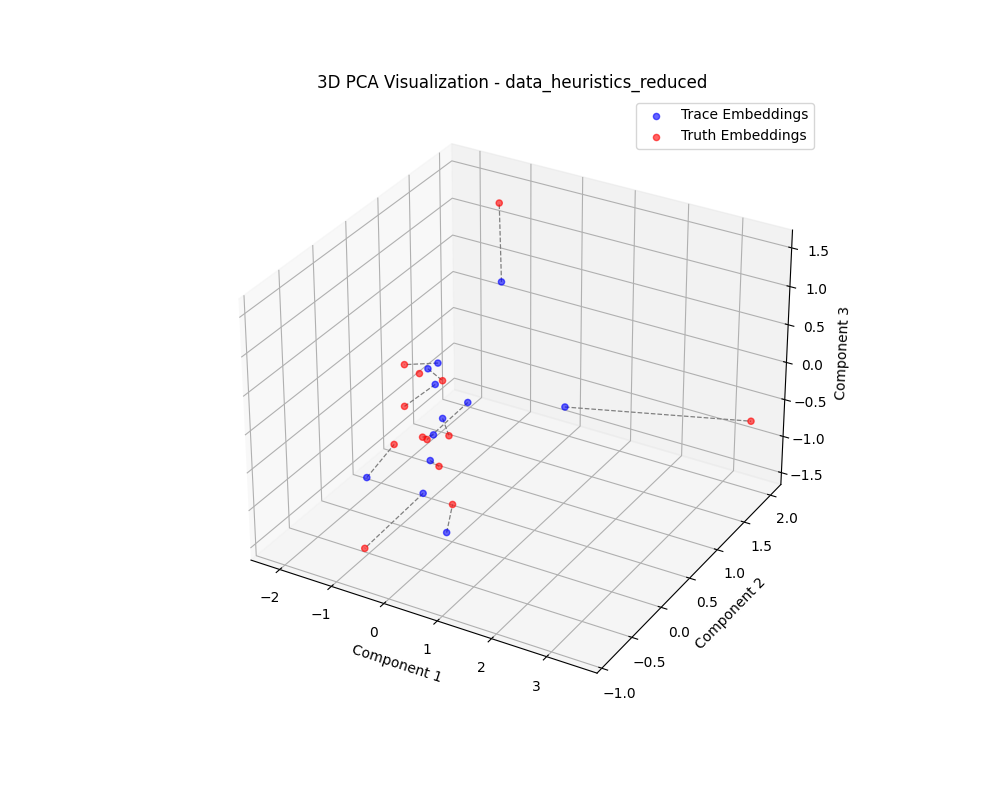}
    \caption{3D components by PCA Heuristics Miner on Road Traffic Fines dataset}
    \label{fig:roadhurisnodes}
\end{figure}

In addition to evaluating the overall performance of the discovered process model, this approach enables the identification of deviations at the activity level. Specifically, this analysis highlights which nodes (activities) diverge from the expected process model and provides insights into the sources of distance observed in the embeddings. Although the vectors are originally embedded in a 128-dimensional space during training, Principal Component Analysis (PCA) was applied to reduce the dimensionality to three for visualization purposes. This enables analysts to identify activities that violate the expected behavior and to interpret compliance deviations in terms of the alignment between observed traces and the event log. Fig. \ref{fig:roadhurisnodes} illustrates the relative distances between activities, allowing us to pinpoint those that exhibit substantial deviations. Such activities may introduce inconsistencies into the analysis. In cases where these deviations correspond to outdated or irrelevant activities, they can be selectively filtered to improve the robustness of subsequent analyses.

\begin{figure}[htbp]
    \centering
    \includegraphics[width=\textwidth, height=0.9\textheight, keepaspectratio]{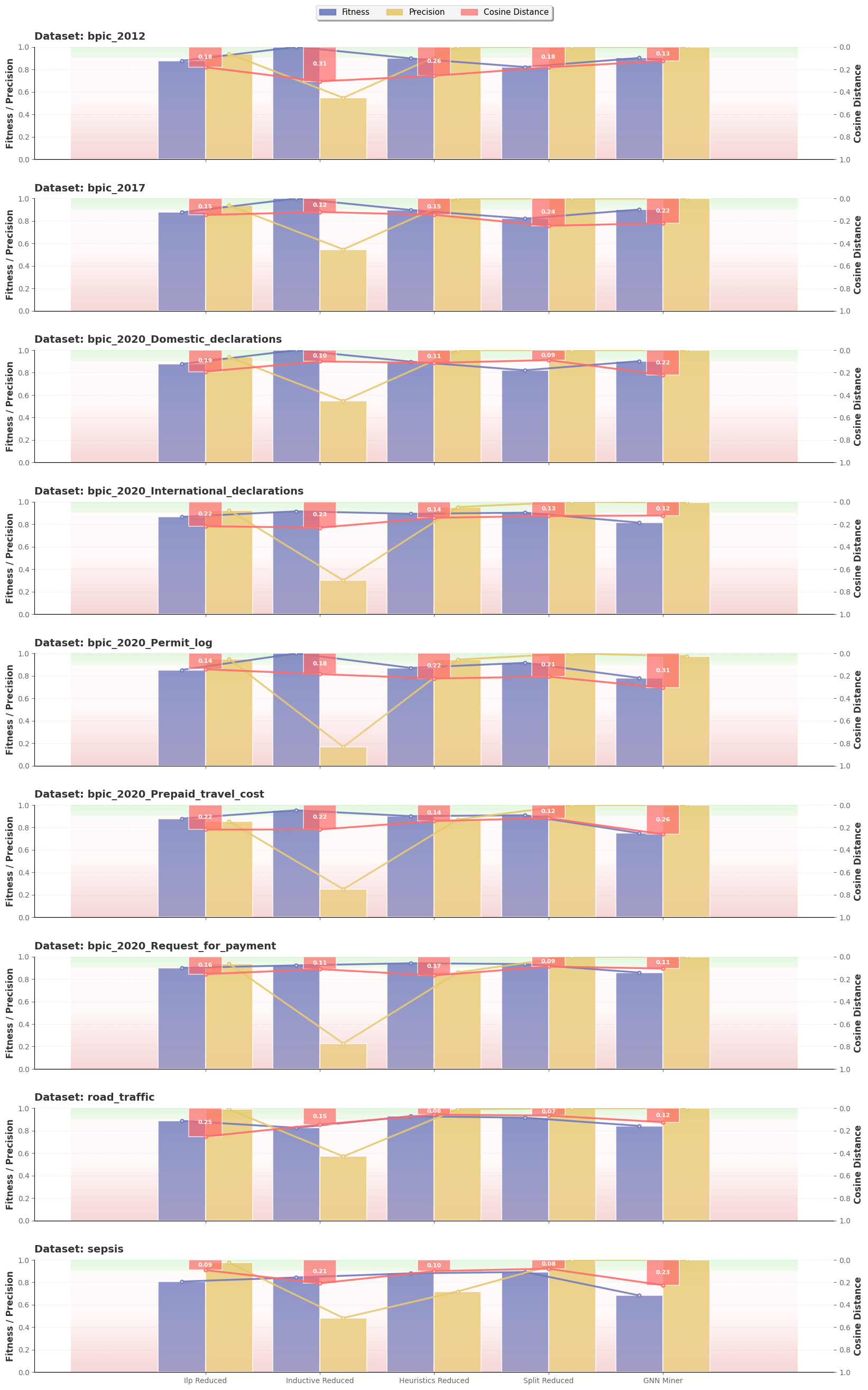}
    \caption{Context-aware process model conformance checking results vs. cosine distance}
    \label{fig:contextbasedcomparison}
\end{figure}

In real-world scenarios, the absence of contextual information often makes it impractical to filter activity sets on a per-case basis. Consequently, the static approach offers a more generalizable and robust framework for comparison. Nevertheless, in situations where the relevant activity set can be explicitly defined, the discovered process model may be filtered accordingly. This enables the computation of the distance between the filtered model and the corresponding ground truth, thereby allowing for a more targeted assessment of structural similarity. Fig. \ref{fig:contextbasedcomparison} presents the embedding-based cosine distance along with fitness and precision scores for context-aware models.

\begin{figure}[htbp]
    \centering
    \begin{subfigure}[b]{0.38\linewidth}
        \centering
        \includegraphics[width=\linewidth]{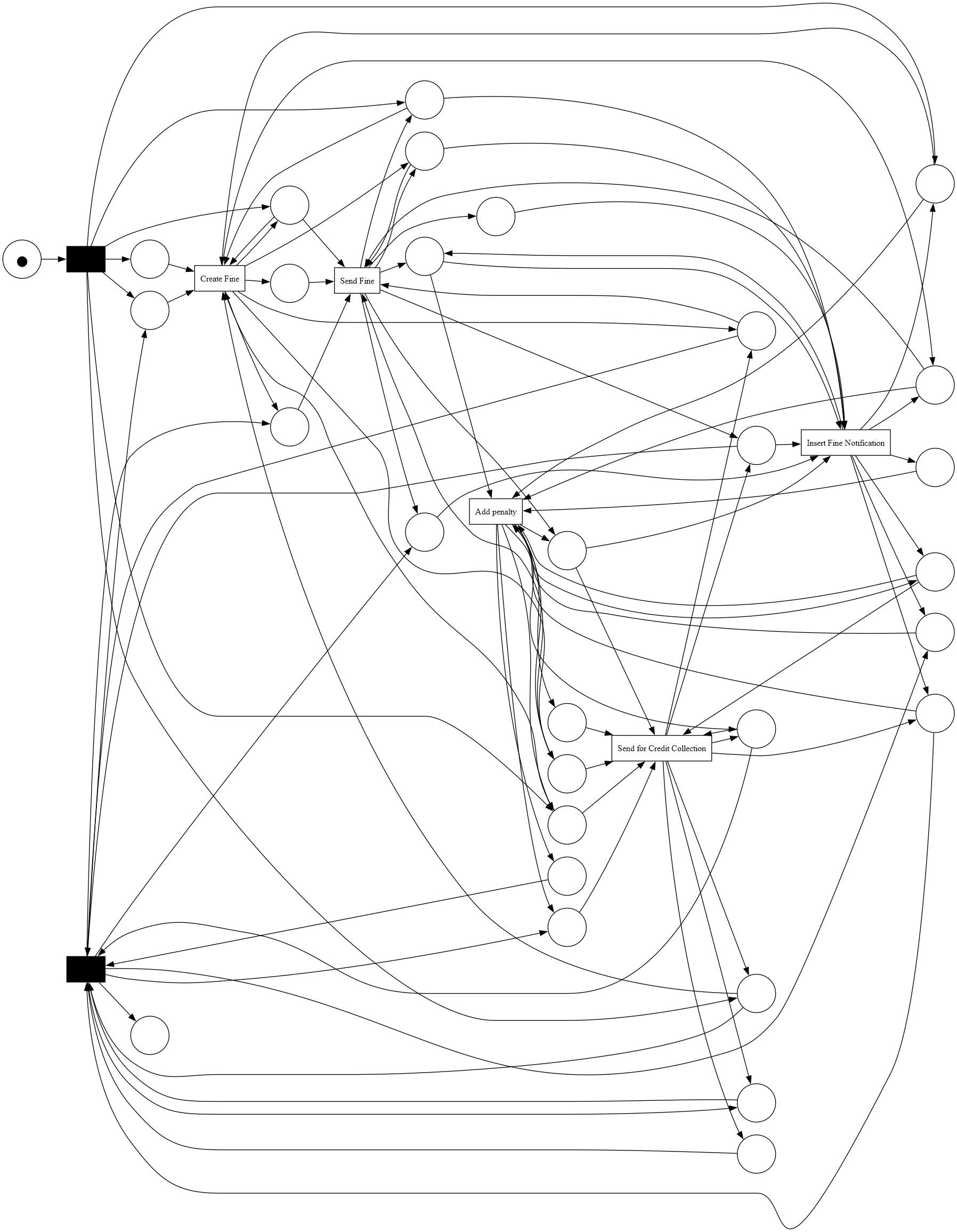}
        \caption{ILP miner}
        \label{fig:roadilp}
    \end{subfigure}

    \vspace{6pt}
    \begin{subfigure}[b]{\linewidth}
        \centering
        \includegraphics[width=\linewidth]{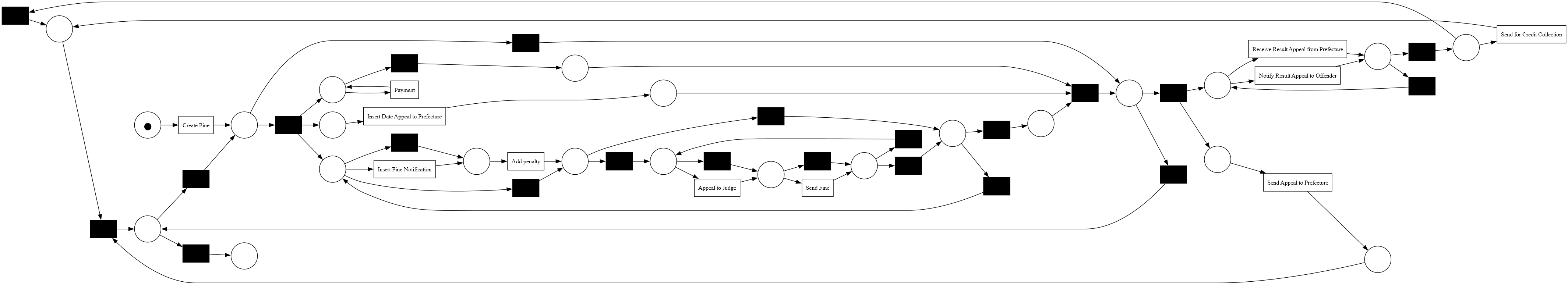}
        \caption{Inductive Miner}
        \label{fig:roa}
    \end{subfigure}
    \caption{Discovered process model for Road Traffic Fines dataset}
    \label{fig:road_comparison}
\end{figure}

Similar to the static model, Fig. \ref{fig:contextbasedcomparison} demonstrates that cosine distance generally follows the same trend as fitness and precision. Nevertheless, certain discrepancies emerge relative to the expected outcomes. For instance, in the Road Traffic Fine and BPIC\_2020\_Request\_for\_Payment datasets, the ILP Miner yields higher distances than the Inductive Miner, despite simultaneously achieving higher fitness and precision. This inconsistency can be further examined by visualizing the discovered process models of both methods (Fig. \ref{fig:road_comparison}). The ILP miner, by design, covers a wide range of possible scenarios. While this increases fitness and precision, important indicators for performance evaluation, it does so at the cost of simplicity. In practice, this means that graph embedding methods, such as random walks, must traverse a larger number of potential edges. As the number of edges grows, the distance between the ground truth and the discovered process model also increases. Thus, fitness and precision alone cannot fully capture model quality. Other indicators, such as generalizability and simplicity, are also important for assessing conformance \cite{buijs2014quality}. This explains why cosine distance can serve as a more balanced criterion in certain cases. The distinction between static and context-aware settings further clarifies this behavior. In the context-aware case, when activities are filtered based on $\sigma$, the corresponding edges remain connected, which can lead to the inclusion of all possible behavioral variants. This often produces the so-called flower model, characterized by excessive generalization \cite{van2011representational}. By contrast, in the static model, the number of activities is fixed, and the random walk is constrained by limited walk lengths. These restrictions prevent chaotic walk behavior and enable a more meaningful comparison to the ground truth behavior ($\mathcal{B}_{L}$).

For instance, in the Road Traffic and BPIC\_2020\_Request\_for\_Payment datasets, the simplicity metric demonstrates that our approach also accounts for process model complexity, an essential dimension of conformance checking. While increasing the number of edges can artificially improve fitness, it simultaneously reduces the readability and interpretability of the process model. In other words, overly complex models can mislead evaluation by interrupting the balance between accuracy and comprehensibility.

Similarly, the gap observed between the Heuristics Miner and the Split Miner on the BPIC\_2012 dataset, where the simplicity metrics in Tab. \ref{tab:simplicity-indicators} differ by approximately 20\%, further illustrates how traditional conformance measures may overlook structural and behavioral aspects that are effectively captured through distance-based analysis.

\begin{table}[!htb]
\centering
\caption{Simplicity metrics for process models discovered using different methods across datasets. Higher values indicate more readable models.
\textit{Note:} Simplicity metric as defined in \citep{sommers2023supervised}.}
\label{tab:simplicity-indicators}
\begin{tabular}{@{}llc@{}}
\toprule
\textbf{Dataset} & \textbf{Method} & \textbf{Simplicity} \\
\midrule
BPIC\_2020\_Request\_for\_Payment & ILP Miner & 0.245 \\
           & Inductive Miner & 0.556 \\
\addlinespace
Road Traffic & ILP Miner & 0.245 \\
             & Inductive Miner & 0.636 \\
\addlinespace
BPIC\_2012 & Heuristics Miner & 0.654 \\
           & Split Miner & 0.818 \\
\bottomrule
\end{tabular}
\end{table}

In addition to Fig.~\ref{fig:minedcomp} and Fig.~\ref{fig:contextbasedcomparison}, Appendix~\ref{comp2} presents an alternative visualization of the relationship between fitness, precision, and cosine distance, further confirming the alignment between these two categories of metrics.

\subsection{Unmatched nodes}

Another source of higher distance in some cases is the unmatched nodes. In highly complex process models particularly methods such as ILP miner, the number of unmatched nodes (nodes present in the event log but absent in the discovered model) tends to increase, as it becomes less likely for the discovered model to visit all relevant nodes in $\mathcal{M}$ in the random walk. Although we employed a dynamic hyperparameter selection strategy to accommodate varying model complexities, full node coverage was not always achievable, depending on the process discovery algorithm used. This limitation leads to inflated distance values for certain models. Conversely, when the distances are reduced under the static approach, this may indicate the presence of simpler yet structurally comprehensive models, capable of capturing a wide range of behaviors with fewer deviations.

\begin{figure}[t]
    \centering
    \includegraphics[width=0.5\linewidth]{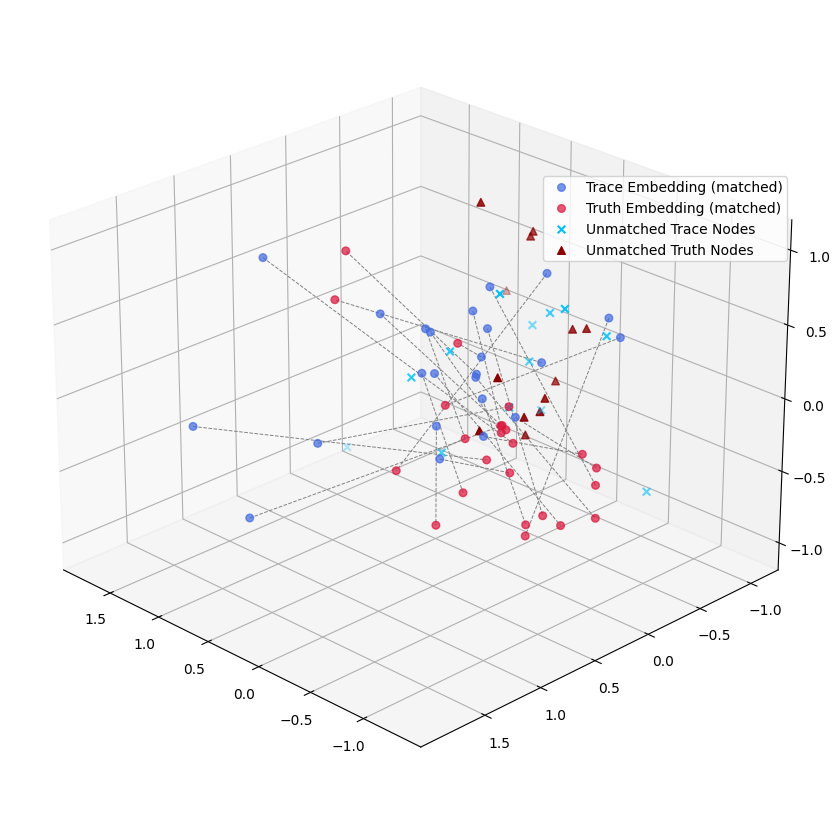}
    \caption{3D Components by PCA ILP miner on 2020 Domestic Declaration  }
    \label{fig:ilppca}
\end{figure}

In computing the distance, the maximum distance among the matched nodes is assigned to the unmatched nodes. This approach ensures that the measure more accurately reflects behavioral alignment, offering a balanced alternative to assigning an arbitrary value such as zero or the maximum cosine distance (\textit{i.e., 2}). The presence of numerous unmatched nodes, which negatively affects both the cosine distance and fitness. As shown in Fig. \ref{fig:ilppca}, which visualizes the principal component analysis (PCA) of node embeddings for the ``2020 Domestic Declaration'' dataset under the ILP miner, more than half of the nodes are either missing or unmatched (with ground truth embeddings shown in the figure).

In contrast, the Inductive Miner (IM) demonstrates high fitness and performs well in terms of cosine distance. Despite its relatively low precision compared to other methods, its cosine distance remains closely aligned with the ground truth. This suggests that graph embedding-based distance metrics offer a balanced representation that reflects both fitness and precision. Importantly, unmatched nodes are not compared through aligned cosine distances. Instead, they receive the penalty term directly, because no valid node-to-node correspondence is available for geometric comparison. As a result, the underconstrained nature of the Procrustes solution when \( |\mathcal{S}| \) is small does not directly affect the distances assigned to unmatched nodes. For the matched anchors, the total and per-anchor Procrustes residuals reported in Tab.~\ref{tab:dataset-method-sweeps-context} and Tab.~\ref{tab:dataset-method-sweeps-static} provide a separate diagnostic of alignment quality across dataset-method combinations.

\subsection{Robustness of model-centric evaluation}

As outlined in our methodology, we implemented a node embedding approach combined with a distance function to decompose process models into localized neighborhoods using a random walk strategy with dynamic walk lengths. This design captures long-term dependencies between activities and embeds the structural information through the skip-gram algorithm. The objective is to detect deviations and errors, including both behavioral outliers and event recording anomalies. An essential expectation is that, as deviations in the event log increase, the evaluation criteria should remain stable as long as the reference process model is fixed. This property is particularly critical for achieving robust evaluation under noisy conditions.

\begin{figure}[ht]
    \centering
    \begin{subfigure}{0.49\linewidth}
        \centering
        \includegraphics[width=\linewidth]{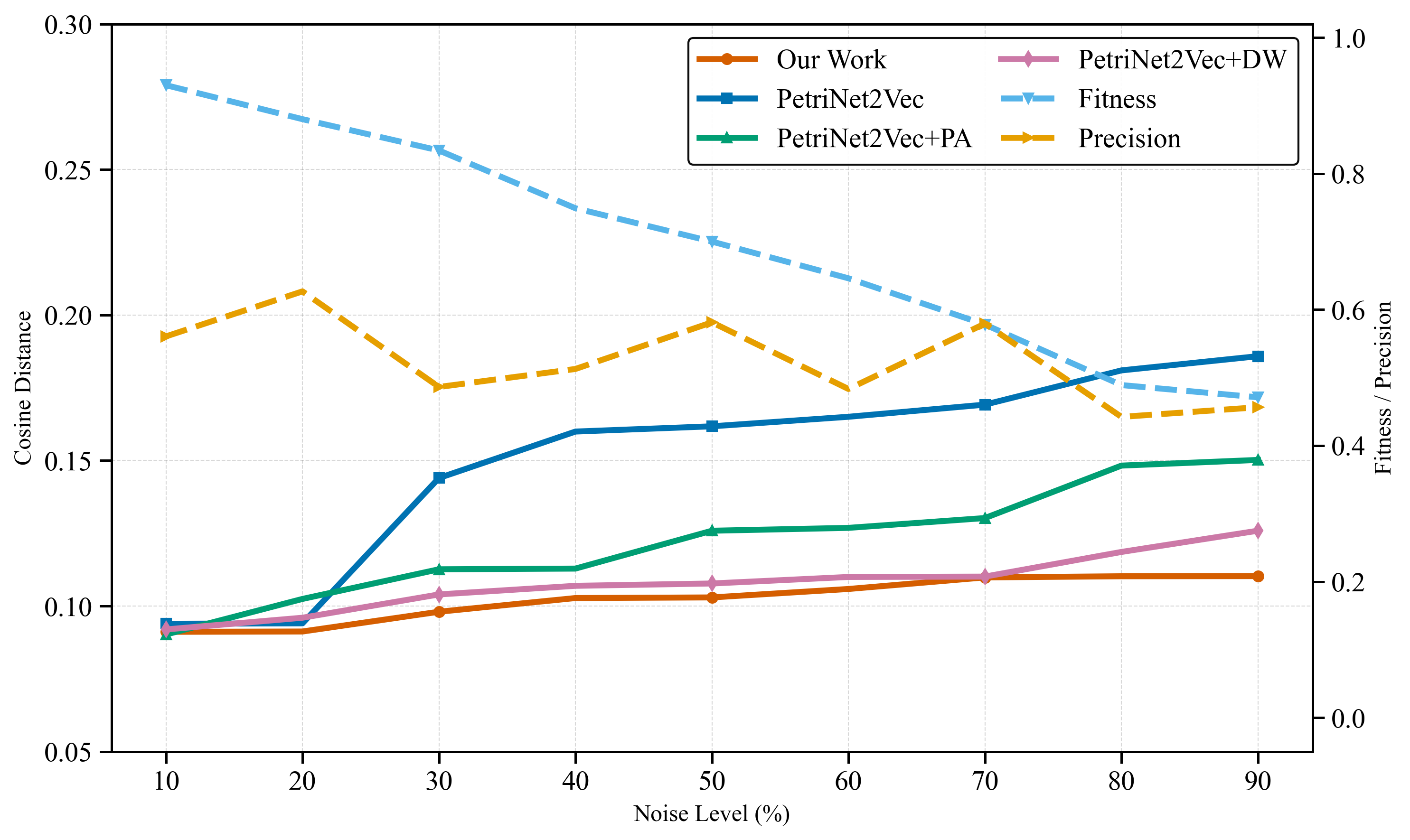}
        \caption{Evaluation of $\mathcal{M}_{0}$ on event logs with increasing noise levels ($\mathcal{L}_\varepsilon$).}
        \label{fig:noiseinc}
    \end{subfigure}
    \hfill
    \begin{subfigure}{0.49\linewidth}
        \centering
        \includegraphics[width=\linewidth]{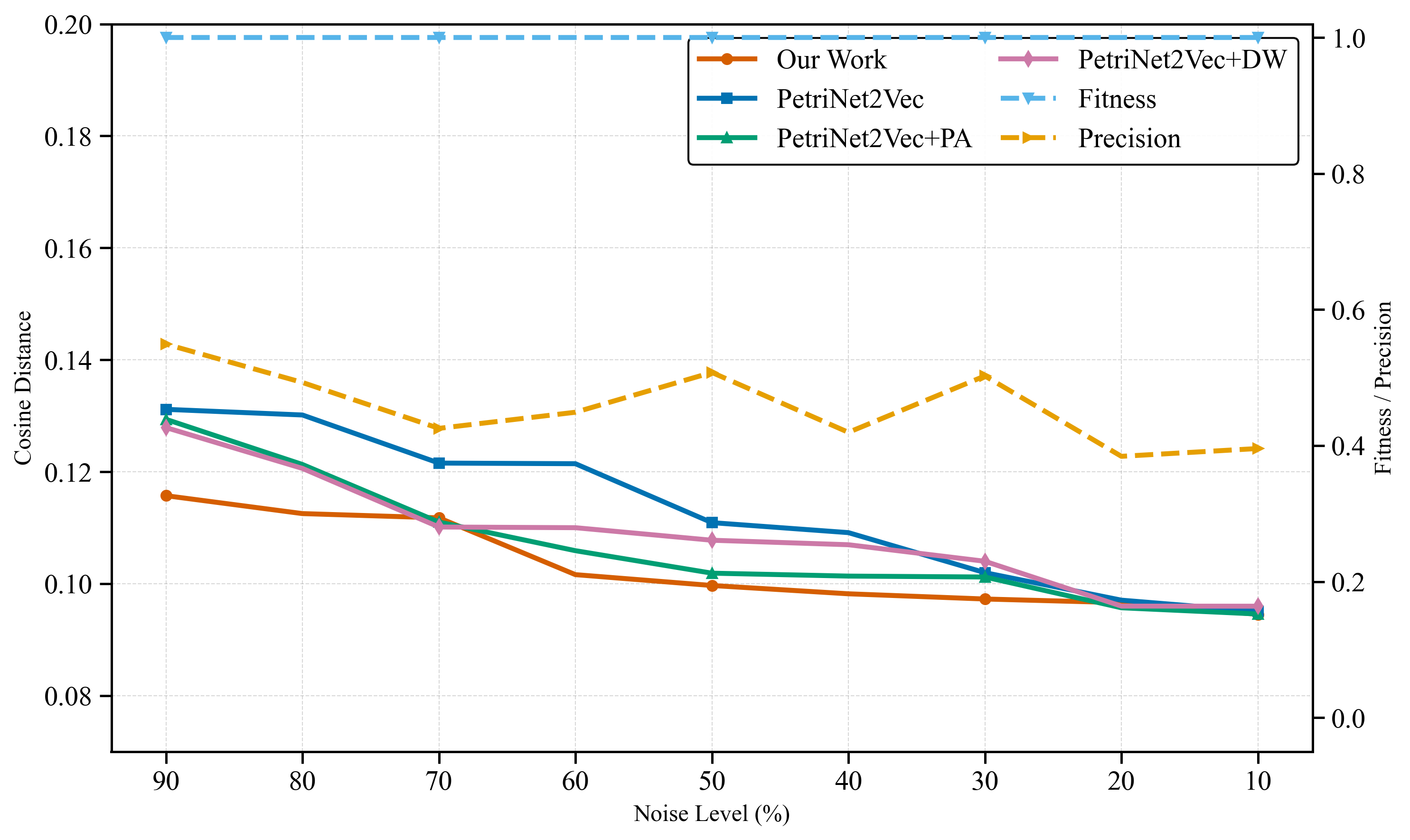}
        \caption{Evaluation of $\mathcal{M}_{90}$ on event logs with decreasing noise levels ($\mathcal{L}_\varepsilon$).}
        \label{fig:noisedec}
    \end{subfigure}
    
    \caption{Distance and conformance metrics evaluated on fixed models across varying noise levels in the event logs.}
    \label{fig:noise_combined}
\end{figure}

To evaluate this, we employed the synthetic simulator described in Section \ref{exp}. We first generated a base event log $\mathcal{L}_0$ aligned with a sound process model $\mathcal{M}_0$, and then incrementally introduced noise to obtain a series of logs $\mathcal{L}_\varepsilon$ for $\varepsilon \in \{10\%, 20\%, \dots, 90\%\}$. We compared our proposed method against alignment-based conformance checking metrics (fitness and precision) as well as the recent task embedding approach, PetriNet2Vec \citep{COLONNA2024200423}. In PetriNet2Vec, embeddings are learned for transition pairs using a fixed context window, and similarity is computed across separate embedding spaces. To enable a more comprehensive comparison, we extended PetriNet2Vec in two ways: (1) PetriNet2Vec with Dynamic Window (PetriNet2Vec+DW), which introduces a variable context window proportional to the random walk length, and (2) PetriNet2Vec with Procrustes Alignment (PetriNet2Vec+PA), which aligns embeddings into a shared vector space. These variants allow us to isolate and assess the individual contributions of dynamic context modeling and embedding alignment, as well as their combined effect in our proposed method. As shown in Fig.~\ref{fig:noiseinc}, our proposed node embedding approach (red curve) maintains a remarkably stable cosine distance across increasing noise levels. The curve exhibits only minor fluctuations, with a slight upward trend under extreme noise conditions. The right axis corresponds to alignment-based conformance checking results, shown with dashed lines, while the left axis represents distance-based metrics, depicted with solid lines. In contrast, the benchmark methods such as PetriNet2Vec (blue), fitness (light blue, dashed), and precision (yellow, dashed) demonstrate pronounced degradation beyond 30\% noise. As indicated, both fitness and precision values drop sharply, and the cosine distance of PetriNet2Vec increases substantially. This behavior highlights the inability of the unchanged model $\mathcal{M}_{0}$ to reproduce increasingly noisy behavior, leading to either underfitting (low fitness) or overgeneralization (low precision).

The two intermediate variants, PetriNet2Vec+PA (green) and PetriNet2Vec+DW (pink), further clarify the contributions of alignment and dynamic context modeling. As illustrated, introducing Procrustes Alignment (PA) consistently reduces variability compared to the baseline PetriNet2Vec, indicating the benefit of embedding alignment in producing comparable vector spaces. Meanwhile, the Dynamic Window (DW) configuration closely follows the trend of our method, confirming its critical role in capturing both local and global dependencies during training. As reported in Tab.~\ref{tab:merged_cv_metrics}, incorporating the dynamic window (DW) alone yields a notable stability improvement, reflected by lower coefficients of variation (CV). When both mechanisms are combined in our proposed framework, the performance further improves by 22.83\% over the strongest variant (PetriNet2Vec+DW), achieving the highest robustness across all noise levels.

Our node embedding-based approach preserves the local neighborhood structure of nodes by executing biased random walks with a dynamic window and training a skip-gram model on the resulting node sequences. In the context of process mining, this means that each activity’s embedding is primarily shaped by its surrounding transitions (the activities that typically precede or follow it) while also capturing long-term dependencies. Consequently, minor process deviations or noise do not substantially disrupt these local neighborhoods. Even under high noise levels (up to 90\%), the dominant directly-follows relations often remain intact, allowing the method to anchor activity embeddings to the underlying process structure.

In a complementary experiment, we evaluated the performance of the model $\mathcal{M}_{90}$, discovered from a highly noisy log $\mathcal{L}_{90}$, when applied to logs with progressively decreasing noise levels. From a theoretical standpoint, fitness is expected to remain stable as noise decreases, since a model that generalizes from noisy data will typically accept all valid traces contained in cleaner logs. In contrast, precision is expected to deteriorate in this setting, as the model continues to permit execution paths that no longer appear in the less noisy event logs. This outcome reflects the inherent tendency of models discovered from highly noisy logs to overgeneralize.

As illustrated in Fig.~\ref{fig:noisedec}, decreasing noise leads to a pronounced decline in both precision and PetriNet2Vec performance, whereas our node embedding approach exhibits only minor fluctuations. Interestingly, PetriNet2Vec shows less variability in this setting because $\mathcal{M}_{90}$, derived from the most noisy log, already encodes a broad range of possible traces; this extensive coverage reduces the observable change when compared to the ground truth. The extended variants (\textit{i.e.,} PetriNet2Vec+PA and PetriNet2Vec+DW) demonstrate similar trends, performing better than the baseline PetriNet2Vec. Consistent with the previous experiment, the dynamic window variant shows behavior closely aligned with our proposed method, confirming its crucial role in capturing both local and global dependencies. The coefficients of variation, reported in Tab.~\ref{tab:merged_cv_metrics}, further reinforce these observations: precision exhibits the highest instability under decreasing noise, while the embedding-based methods maintain significantly greater robustness. Notably, our approach outperforms PetriNet2Vec+DW by 18.78\%, maintaining lower variation and demonstrating superior capability in preserving stable, localized structural properties of the process. This contrasts with precision, which remains more sensitive to global behavioral deviations.

\begin{table}[!htb]
\centering
\setlength{\tabcolsep}{8pt}
\caption{Comparison of coefficient of variation (CV) values under different noise conditions in $\mathcal{L}_\varepsilon$ and $\mathcal{M}_\varepsilon$. \\{\footnotesize (PA: Procrustes Alignment, DW: Dynamic Window)}}
\label{tab:merged_cv_metrics}
\begin{tabular}{@{}lcc@{\hspace{28pt}}c@{}}
\toprule
\textbf{Metric} & \textbf{CV$@$\textit{($\varepsilon \downarrow$ in $\mathcal{L}_{\varepsilon}$)}} & \textbf{CV$@$\textit{($\varepsilon \uparrow$ in $\mathcal{L}_{\varepsilon}$)}} & \textbf{CV$@$\textit{($\varepsilon \uparrow$ in $\mathcal{M}_{\varepsilon}$)}} \\
\midrule
Fitness        & 0.000000 & 0.237964 & 0.000000 \\
Precision      & 0.124164 & 0.121690 & 0.126832 \\
PetriNet2Vec \citep{COLONNA2024200423}   & 0.120372 & 0.227511 & 0.105085 \\
PetriNet2Vec+PA   & 0.109452 & 0.162488 & 0.256595 \\
PetriNet2Vec+DW  & 0.095438 & 0.096008 & 0.524864 \\
\textbf{Our Work} & \textbf{0.077518} & \textbf{0.074094} & \textbf{0.678094} \\
\midrule
$\Delta$ {\footnotesize\textit{(Improvement over baseline \%)}} & \textit{18.78} & \textit{22.83} & \textit{29.19} \\
\bottomrule
\end{tabular}
\end{table}

In general, our goal in conformance checking is to identify an evaluation metric that assesses the quality of a process model independently of the specific event log used for discovery. This allows for a fair comparison between both automated and manually constructed models, without being confounded by deviations or noise present in the event log. While such an approach cannot eliminate the impact of noise entirely, it can significantly reduce its influence on the evaluation. As demonstrated in Fig. \ref{fig:noise_combined}, our method exhibits a degree of tolerance to noise, effectively capturing structural differences even when logs contain behavioral anomalies. This highlights the potential of embedding-based methods to provide stable and model-centric evaluation across a variety of process discovery scenarios. 

\subsection{Assessment of deviated process models}

In all previous experiments, the ground truth event log $\mathcal{L}$ was affected by varying levels of noise, which in turn influenced the discovered models. In this experiment, the event log is kept clean and fixed as $\mathcal{L}_0$, while the discovered process models $\mathcal{M}_\varepsilon$ are derived from logs with increasing levels of noise. This setup isolates the impact of model degradation, enabling an analysis of how embedding-based distances reflect structural deviations in the discovered models. By holding the event log constant, the experiment focuses solely on evaluating the sensitivity of our method to model corruption.

\begin{figure}[htbp]
    \centering
    \includegraphics[width=0.51\linewidth]{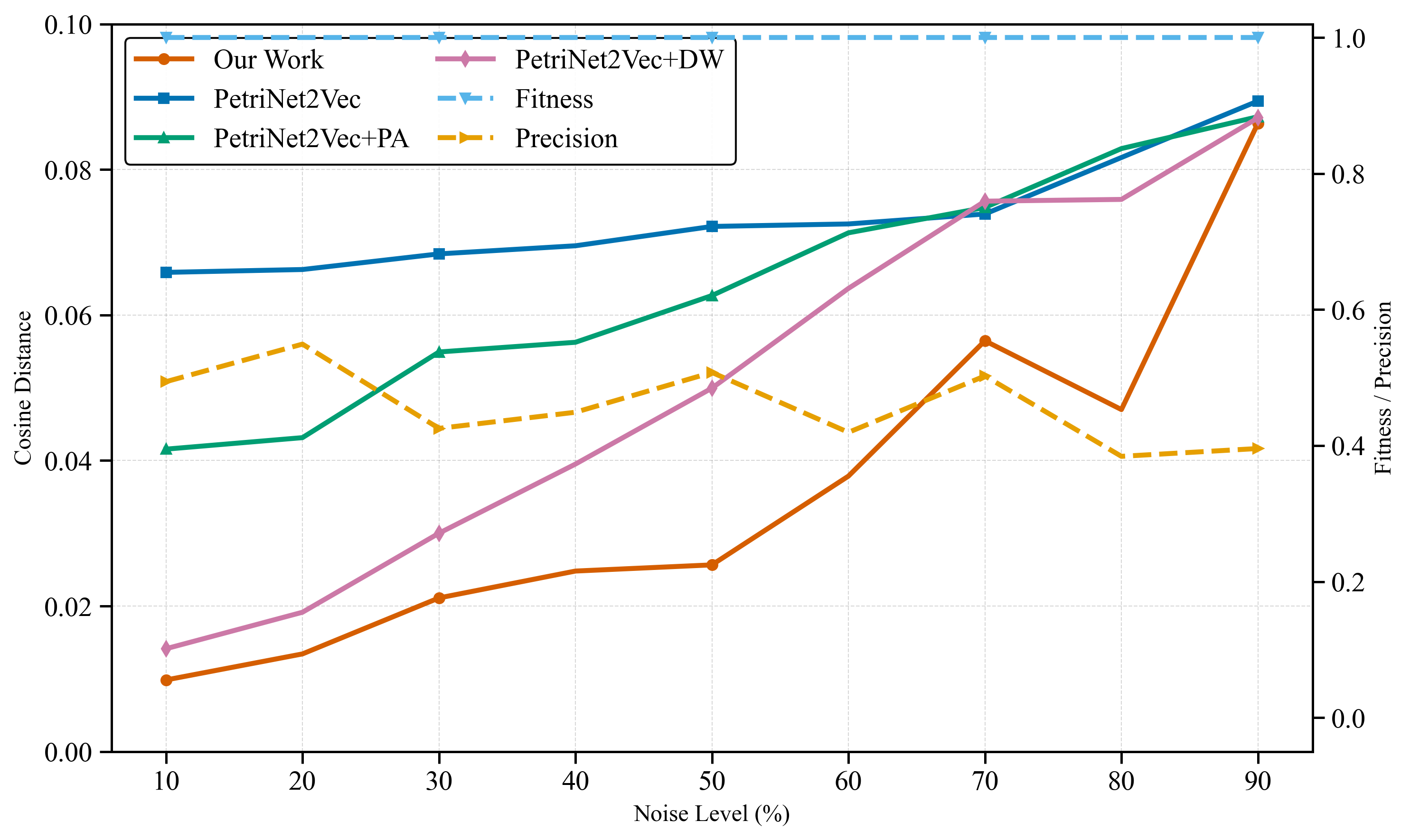}
    \caption{Evaluation of distance and conformance metrics using noise-free event logs with process models discovered under varying noise levels. ($\mathcal{M}_\varepsilon$)}
    \label{fig:combined_distance_corruptedmodel}
\end{figure}

As shown in Fig.~\ref{fig:combined_distance_corruptedmodel}, fitness remains relatively unchanged across varying noise levels, failing to capture the increasing divergence between the discovered models and the true process behavior. This confirms fitness’s insensitivity to model corruption when the event log itself is noise-free. In contrast, our proposed approach exhibits a clear and monotonic increase with higher model noise, effectively reflecting the progressive structural degradation of the process model. While precision, PetriNet2Vec, and their variants also deteriorate with increasing noise, their responses are less consistent and less sensitive compared to our embedding-based approach. As reported in Tab.~\ref{tab:merged_cv_metrics}, the higher coefficients of variation (CV) indicate that PetriNet2Vec is somewhat more sensitive than precision; however, its variants exhibit notable improvements. Our proposed method demonstrates a 29.19\% higher CV compared to the best-performing PetriNet2Vec variant (with a dynamic window), indicating that the proposed distance metric is more responsive to structural discrepancies across corrupted models. At the same time, it remains robust to log noise, making it a reliable indicator for evaluating process model quality independently of event log anomalies. In this context, the F1-score may be viewed as a combined summary of fitness and precision, whose trend is already determined by the behavior of these two metrics and therefore does not require separate reporting in the robustness analysis. Since the F1-score is the harmonic mean of fitness and precision and is monotonically increasing in both, any decrease in one metric while the other remains stable will necessarily lead to a corresponding decrease in F1, often even more noticeably due to the sensitivity of the harmonic mean to lower values (Fig.~\ref{fig:noise_combined}). Conversely, when both fitness and precision remain stable, the F1-score also remains stable. For this reason, reporting F1 separately would largely reproduce the same qualitative pattern already conveyed by the individual fitness and precision results.

\section{Conclusions, Future Directions, and Limitations}
\label{conclusion}

In this work, we proposed a node embedding-based distance framework for evaluating process discovery. The approach comprises four key stages: (1) a PetriNetGraph constructor that systematically transforms discovered models into graph structures suitable for graph representation learning; (2) a process-aware directed random walk strategy designed to capture both short- and long-term activity dependencies; (3) embedding of both the discovered and ground-truth models followed by Procrustes alignment to ensure comparability in a shared vector space; and (4) an entropy-regularized distance aggregation function that provides a unified, bounded measure of structural similarity.

We benchmarked the proposed framework against alignment-based conformance metrics (fitness and precision) and the recent task embedding method, PetriNet2Vec, along with two intermediate variants designed to isolate the effects of the dynamic window and Procrustes alignment mechanisms. The results demonstrate that our approach consistently delivers reliable performance across real-life datasets and remains robust under noisy conditions, as validated through controlled experiments on synthetic logs with incremental noise levels. Unlike fitness and precision (which degrade rapidly in the presence of noise in the event log) and PetriNet2Vec, our method maintains stability by leveraging random walks to capture long-term dependencies and smooth out local perturbations. The robustness of the proposed approach is further confirmed by the coefficients of variation, showing 18.78\% and 22.83\% improvements under decreasing and increasing event-log noise, respectively, and a 29.19\% higher CV in process-model noise experiments, indicating superior sensitivity to structural deviations while preserving resistance to log anomalies. We showed that the proposed approach not only provides a reliable metric reflecting the behavior of traditional measures such as fitness and precision, but also enables the identification of deviations and outliers across activities in both the discovered process model and the ground truth. This capability extends its utility to compliance checking, which is highly beneficial for performance analysis and process enhancement.

The findings highlight that embedding-based structural distances offer a principled and event-log–agnostic complement to traditional conformance metrics for process model evaluation. Beyond providing a more resilient metric, this work opens new opportunities for integrating AI-driven optimization into process discovery, for instance, by employing embedding-based distances as loss functions to guide model training and improve robustness. Although the proposed framework has been comprehensively evaluated across multiple benchmark event logs and experimental scenarios to ensure consistency with the study’s objectives, its application potential extends beyond the process mining domain. The entropy-regularized node aggregation and Procrustes-based alignment are not confined to process models; they can be adapted for a broader range of graph representation tasks involving complex relational structures. The combination of the Procrustes alignment technique with the skip-gram model further opens an avenue for cross-graph embeddings, enabling structural comparison not only in process mining but also in domains such as supply chain networks, social networks, and other relational systems. Future research could explore these directions to assess the generalizability and scalability of the proposed approach under diverse graph topologies and domain-specific characteristics.

\bibliographystyle{splncs04} 
\bibliography{Ref}

\clearpage
\subsection*{Appendix}

\appendix
\renewcommand{\theHsection}{appendix.\Alph{section}}
\section{Extended Results of $\lambda$ Variations Under Context-Aware and Static Settings}
\label{lambda}

Tab. \ref{tab:dataset-method-sweeps-context} presents different datasets and methods with varying $\lambda$ in the context-aware scenario, and reports the number of matched and unmatched nodes between the truth and the trace. As expected, the distance increases with higher values of $\lambda$, as the weighting scheme places less emphasis on smaller distances. We additionally report the total Procrustes residual \( \|\mathbf{E}_A \mathbf{W} - \mathbf{E}_B\|_F \) and the per-anchor residual \( \|\mathbf{E}_A \mathbf{W} - \mathbf{E}_B\|_F / |\mathcal{S}| \) to quantify the geometric fit of the alignment step over the shared anchors and to distinguish alignment error from the penalty assigned to absent nodes.

\begingroup
\small
\setlength{\tabcolsep}{0pt}
\setlength{\LTleft}{0pt}
\setlength{\LTright}{0pt plus 1fil}
\renewcommand{\arraystretch}{1.08}
\begin{longtable}{@{\extracolsep{\fill}}l c c c c c c@{}}
\caption{Context-aware model distance of node embeddings, grouped by dataset and method, with varying $\lambda$.}
\label{tab:dataset-method-sweeps-context}\\
\toprule
Dataset / Method & $\lambda$ & $D_{\text{Weighted}}$ & \# Matched & \# Unmatched & Total residual & Per-anchor \\
\midrule
\endfirsthead
\multicolumn{7}{@{}l}{\textbf{\tablename~\thetable}\ (continued)}\\[2pt]
\toprule
Dataset / Method & $\lambda$ & $D_{\text{Weighted}}$ & \# Matched & \# Unmatched & Total residual & Per-anchor \\
\midrule
\endhead
\midrule
\multicolumn{7}{r@{}}{\footnotesize\itshape Continued on next page}\\
\endfoot
\bottomrule
\endlastfoot

\multicolumn{7}{@{}l}{\textbf{bpic\_2012}} \\*
\quad GNN Miner          & 0   & 0.00766 & 12 & 14 & 6.206 & 0.517 \\*
                         & 0.5 & 0.11635 & 12 & 14 & 6.206 & 0.517 \\*
                         & 1   & 0.12079 & 12 & 14 & 6.206 & 0.517 \\*
                         & 10  & 0.12468 & 12 & 14 & 6.206 & 0.517 \\*
                         & inf & 0.12510 & 12 & 14 & 6.206 & 0.517 \\*
\addlinespace[3pt]
\quad Heuristics Miner   & 0   & 0.01317 & 12 & 14 & 5.343 & 0.445 \\*
                         & 0.5 & 0.21514 & 12 & 14 & 5.343 & 0.445 \\*
                         & 1   & 0.23656 & 12 & 14 & 5.343 & 0.445 \\*
                         & 10  & 0.25490 & 12 & 14 & 5.343 & 0.445 \\*
                         & inf & 0.25678 & 12 & 14 & 5.343 & 0.445 \\*
\addlinespace[3pt]
\quad ILP Miner          & 0   & 0.02160 & \phantom{0}9 & 17 & 5.613 & 0.624 \\*
                         & 0.5 & 0.16784 & \phantom{0}9 & 17 & 5.613 & 0.624 \\*
                         & 1   & 0.17453 & \phantom{0}9 & 17 & 5.613 & 0.624 \\*
                         & 10  & 0.18016 & \phantom{0}9 & 17 & 5.613 & 0.624 \\*
                         & inf & 0.18075 & \phantom{0}9 & 17 & 5.613 & 0.624 \\*
\addlinespace[3pt]
\quad Inductive Miner    & 0   & 0.02250 & 12 & 14 & 6.207 & 0.517 \\*
                         & 0.5 & 0.22316 & 12 & 14 & 6.207 & 0.517 \\*
                         & 1   & 0.26470 & 12 & 14 & 6.207 & 0.517 \\*
                         & 10  & 0.30227 & 12 & 14 & 6.207 & 0.517 \\*
                         & inf & 0.30632 & 12 & 14 & 6.207 & 0.517 \\*
\addlinespace[3pt]
\quad Split Miner        & 0   & 0.01971 & 12 & 14 & 6.186 & 0.516 \\*
                         & 0.5 & 0.16382 & 12 & 14 & 6.186 & 0.516 \\*
                         & 1   & 0.17301 & 12 & 14 & 6.186 & 0.516 \\*
                         & 10  & 0.18086 & 12 & 14 & 6.186 & 0.516 \\*
                         & inf & 0.18176 & 12 & 14 & 6.186 & 0.516 \\
\addlinespace[8pt]

\multicolumn{7}{@{}l}{\textbf{bpic\_2017}} \\*
\quad GNN Miner          & 0   & 0.00160 & 10 & 18 & 5.434 & 0.543 \\*
                         & 0.5 & 0.18817 & 10 & 18 & 5.434 & 0.543 \\*
                         & 1   & 0.20397 & 10 & 18 & 5.434 & 0.543 \\*
                         & 10  & 0.21761 & 10 & 18 & 5.434 & 0.543 \\*
                         & inf & 0.21903 & 10 & 18 & 5.434 & 0.543 \\*
\addlinespace[3pt]
\quad Heuristics Miner   & 0   & 0.00868 & 10 & 18 & 3.908 & 0.391 \\*
                         & 0.5 & 0.13504 & 10 & 18 & 3.908 & 0.391 \\*
                         & 1   & 0.14025 & 10 & 18 & 3.908 & 0.391 \\*
                         & 10  & 0.14480 & 10 & 18 & 3.908 & 0.391 \\*
                         & inf & 0.14528 & 10 & 18 & 3.908 & 0.391 \\*
\addlinespace[3pt]
\quad ILP Miner          & 0   & 0.01286 & \phantom{0}7 & 21 & 3.923 & 0.560 \\*
                         & 0.5 & 0.14405 & \phantom{0}7 & 21 & 3.923 & 0.560 \\*
                         & 1   & 0.14535 & \phantom{0}7 & 21 & 3.923 & 0.560 \\*
                         & 10  & 0.14646 & \phantom{0}7 & 21 & 3.923 & 0.560 \\*
                         & inf & 0.14657 & \phantom{0}7 & 21 & 3.923 & 0.560 \\*
\addlinespace[3pt]
\quad Inductive Miner    & 0   & 0.00397 & 10 & 18 & 4.058 & 0.406 \\*
                         & 0.5 & 0.11409 & 10 & 18 & 4.058 & 0.406 \\*
                         & 1   & 0.11688 & 10 & 18 & 4.058 & 0.406 \\*
                         & 10  & 0.11931 & 10 & 18 & 4.058 & 0.406 \\*
                         & inf & 0.11957 & 10 & 18 & 4.058 & 0.406 \\*
\addlinespace[3pt]
\quad Split Miner        & 0   & 0.02652 & 10 & 18 & 4.818 & 0.482 \\*
                         & 0.5 & 0.21753 & 10 & 18 & 4.818 & 0.482 \\*
                         & 1   & 0.23077 & 10 & 18 & 4.818 & 0.482 \\*
                         & 10  & 0.24167 & 10 & 18 & 4.818 & 0.482 \\*
                         & inf & 0.24285 & 10 & 18 & 4.818 & 0.482 \\
\addlinespace[8pt]

\multicolumn{7}{@{}l}{\textbf{bpic\_2020\_Domestic\_declarations}} \\*
\quad GNN Miner          & 0   & 0.06224 & 16 & \phantom{0}3 & 5.098 & 0.319 \\*
                         & 0.5 & 0.19358 & 16 & \phantom{0}3 & 5.098 & 0.319 \\*
                         & 1   & 0.20717 & 16 & \phantom{0}3 & 5.098 & 0.319 \\*
                         & 10  & 0.21948 & 16 & \phantom{0}3 & 5.098 & 0.319 \\*
                         & inf & 0.22087 & 16 & \phantom{0}3 & 5.098 & 0.319 \\*
\addlinespace[3pt]
\quad Heuristics Miner   & 0   & 0.03151 & 18 & \phantom{0}1 & 4.134 & 0.230 \\*
                         & 0.5 & 0.10271 & 18 & \phantom{0}1 & 4.134 & 0.230 \\*
                         & 1   & 0.10771 & 18 & \phantom{0}1 & 4.134 & 0.230 \\*
                         & 10  & 0.11253 & 18 & \phantom{0}1 & 4.134 & 0.230 \\*
                         & inf & 0.11321 & 18 & \phantom{0}1 & 4.134 & 0.230 \\*
\addlinespace[3pt]
\quad ILP Miner          & 0   & 0.06212 & \phantom{0}7 & 12 & 3.083 & 0.440 \\*
                         & 0.5 & 0.17963 & \phantom{0}7 & 12 & 3.083 & 0.440 \\*
                         & 1   & 0.18536 & \phantom{0}7 & 12 & 3.083 & 0.440 \\*
                         & 10  & 0.19026 & \phantom{0}7 & 12 & 3.083 & 0.440 \\*
                         & inf & 0.19078 & \phantom{0}7 & 12 & 3.083 & 0.440 \\*
\addlinespace[3pt]
\quad Inductive Miner    & 0   & 0.02883 & 17 & \phantom{0}2 & 4.483 & 0.264 \\*
                         & 0.5 & 0.09495 & 17 & \phantom{0}2 & 4.483 & 0.264 \\*
                         & 1   & 0.09779 & 17 & \phantom{0}2 & 4.483 & 0.264 \\*
                         & 10  & 0.10043 & 17 & \phantom{0}2 & 4.483 & 0.264 \\*
                         & inf & 0.10072 & 17 & \phantom{0}2 & 4.483 & 0.264 \\*
\addlinespace[3pt]
\quad Split Miner        & 0   & 0.03220 & 19 & \phantom{0}0 & 3.048 & 0.160 \\*
                         & 0.5 & 0.08539 & 19 & \phantom{0}0 & 3.048 & 0.160 \\*
                         & 1   & 0.08699 & 19 & \phantom{0}0 & 3.048 & 0.160 \\*
                         & 10  & 0.08843 & 19 & \phantom{0}0 & 3.048 & 0.160 \\*
                         & inf & 0.08858 & 19 & \phantom{0}0 & 3.048 & 0.160 \\
\addlinespace[8pt]

\multicolumn{7}{@{}l}{\textbf{bpic\_2020\_International\_declarations}} \\*
\quad GNN Miner          & 0   & 0.01064 & 20 & 16 & 6.188 & 0.309 \\*
                         & 0.5 & 0.11744 & 20 & 16 & 6.188 & 0.309 \\*
                         & 1   & 0.12007 & 20 & 16 & 6.188 & 0.309 \\*
                         & 10  & 0.12245 & 20 & 16 & 6.188 & 0.309 \\*
                         & inf & 0.12270 & 20 & 16 & 6.188 & 0.309 \\*
\addlinespace[3pt]
\quad Heuristics Miner   & 0   & 0.01703 & 20 & 16 & 4.001 & 0.200 \\*
                         & 0.5 & 0.12621 & 20 & 16 & 4.001 & 0.200 \\*
                         & 1   & 0.13465 & 20 & 16 & 4.001 & 0.200 \\*
                         & 10  & 0.14208 & 20 & 16 & 4.001 & 0.200 \\*
                         & inf & 0.14320 & 20 & 16 & 4.001 & 0.200 \\*
\addlinespace[3pt]
\quad ILP Miner          & 0   & 0.01089 & 12 & 24 & 4.936 & 0.411 \\*
                         & 0.5 & 0.20505 & 12 & 24 & 4.936 & 0.411 \\*
                         & 1   & 0.21157 & 12 & 24 & 4.936 & 0.411 \\*
                         & 10  & 0.21723 & 12 & 24 & 4.936 & 0.411 \\*
                         & inf & 0.21780 & 12 & 24 & 4.936 & 0.411 \\*
\addlinespace[3pt]
\quad Inductive Miner    & 0   & 0.01577 & 20 & 16 & 5.368 & 0.268 \\*
                         & 0.5 & 0.19514 & 20 & 16 & 5.368 & 0.268 \\*
                         & 1   & 0.21269 & 20 & 16 & 5.368 & 0.268 \\*
                         & 10  & 0.22892 & 20 & 16 & 5.368 & 0.268 \\*
                         & inf & 0.23041 & 20 & 16 & 5.368 & 0.268 \\*
\addlinespace[3pt]
\quad Split Miner        & 0   & 0.02806 & 20 & 16 & 3.863 & 0.193 \\*
                         & 0.5 & 0.11868 & 20 & 16 & 3.863 & 0.193 \\*
                         & 1   & 0.12214 & 20 & 16 & 3.863 & 0.193 \\*
                         & 10  & 0.12527 & 20 & 16 & 3.863 & 0.193 \\*
                         & inf & 0.12561 & 20 & 16 & 3.863 & 0.193 \\
\addlinespace[8pt]

\multicolumn{7}{@{}l}{\textbf{bpic\_2020\_Permit\_log}} \\*
\quad GNN Miner          & 0   & 0.06509 & 20 & 33 & 6.817 & 0.341 \\*
                         & 0.5 & 0.27617 & 20 & 33 & 6.817 & 0.341 \\*
                         & 1   & 0.29163 & 20 & 33 & 6.817 & 0.341 \\*
                         & 10  & 0.30459 & 20 & 33 & 6.817 & 0.341 \\*
                         & inf & 0.30597 & 20 & 33 & 6.817 & 0.341 \\*
\addlinespace[3pt]
\quad Heuristics Miner   & 0   & 0.02288 & 20 & 33 & 4.771 & 0.239 \\*
                         & 0.5 & 0.20987 & 20 & 33 & 4.771 & 0.239 \\*
                         & 1   & 0.21721 & 20 & 33 & 4.771 & 0.239 \\*
                         & 10  & 0.22331 & 20 & 33 & 4.771 & 0.239 \\*
                         & inf & 0.22396 & 20 & 33 & 4.771 & 0.239 \\*
\addlinespace[3pt]
\quad ILP Miner          & 0   & 0.00260 & 12 & 41 & 4.534 & 0.378 \\*
                         & 0.5 & 0.13588 & 12 & 41 & 4.534 & 0.378 \\*
                         & 1   & 0.13882 & 12 & 41 & 4.534 & 0.378 \\*
                         & 10  & 0.14109 & 12 & 41 & 4.534 & 0.378 \\*
                         & inf & 0.14134 & 12 & 41 & 4.534 & 0.378 \\*
\addlinespace[3pt]
\quad Inductive Miner    & 0   & 0.02801 & 20 & 33 & 5.572 & 0.279 \\*
                         & 0.5 & 0.17796 & 20 & 33 & 5.572 & 0.279 \\*
                         & 1   & 0.18101 & 20 & 33 & 5.572 & 0.279 \\*
                         & 10  & 0.18378 & 20 & 33 & 5.572 & 0.279 \\*
                         & inf & 0.18405 & 20 & 33 & 5.572 & 0.279 \\*
\addlinespace[3pt]
\quad Split Miner        & 0   & 0.02329 & 20 & 33 & 5.866 & 0.293 \\*
                         & 0.5 & 0.19381 & 20 & 33 & 5.866 & 0.293 \\*
                         & 1   & 0.20015 & 20 & 33 & 5.866 & 0.293 \\*
                         & 10  & 0.20614 & 20 & 33 & 5.866 & 0.293 \\*
                         & inf & 0.20677 & 20 & 33 & 5.866 & 0.293 \\
\addlinespace[8pt]

\multicolumn{7}{@{}l}{\textbf{bpic\_2020\_Prepaid\_travel\_cost}} \\*
\quad GNN Miner          & 0   & 0.04606 & 21 & 10 & 6.922 & 0.330 \\*
                         & 0.5 & 0.22451 & 21 & 10 & 6.922 & 0.330 \\*
                         & 1   & 0.24235 & 21 & 10 & 6.922 & 0.330 \\*
                         & 10  & 0.25840 & 21 & 10 & 6.922 & 0.330 \\*
                         & inf & 0.26017 & 21 & 10 & 6.922 & 0.330 \\*
\addlinespace[3pt]
\quad Heuristics Miner   & 0   & 0.04031 & 20 & 11 & 6.717 & 0.336 \\*
                         & 0.5 & 0.13817 & 20 & 11 & 6.717 & 0.336 \\*
                         & 1   & 0.14040 & 20 & 11 & 6.717 & 0.336 \\*
                         & 10  & 0.14242 & 20 & 11 & 6.717 & 0.336 \\*
                         & inf & 0.14263 & 20 & 11 & 6.717 & 0.336 \\*
\addlinespace[3pt]
\quad ILP Miner          & 0   & 0.02618 & 12 & 19 & 4.548 & 0.379 \\*
                         & 0.5 & 0.20518 & 12 & 19 & 4.548 & 0.379 \\*
                         & 1   & 0.21222 & 12 & 19 & 4.548 & 0.379 \\*
                         & 10  & 0.21826 & 12 & 19 & 4.548 & 0.379 \\*
                         & inf & 0.21887 & 12 & 19 & 4.548 & 0.379 \\*
\addlinespace[3pt]
\quad Inductive Miner    & 0   & 0.06434 & 20 & 11 & 6.977 & 0.349 \\*
                         & 0.5 & 0.19583 & 20 & 11 & 6.977 & 0.349 \\*
                         & 1   & 0.20673 & 20 & 11 & 6.977 & 0.349 \\*
                         & 10  & 0.21680 & 20 & 11 & 6.977 & 0.349 \\*
                         & inf & 0.21787 & 20 & 11 & 6.977 & 0.349 \\*
\addlinespace[3pt]
\quad Split Miner        & 0   & 0.01560 & 21 & 10 & 2.685 & 0.128 \\*
                         & 0.5 & 0.10475 & 21 & 10 & 2.685 & 0.128 \\*
                         & 1   & 0.11060 & 21 & 10 & 2.685 & 0.128 \\*
                         & 10  & 0.11573 & 21 & 10 & 2.685 & 0.128 \\*
                         & inf & 0.11653 & 21 & 10 & 2.685 & 0.128 \\
\addlinespace[8pt]

\multicolumn{7}{@{}l}{\textbf{bpic\_2020\_Request\_for\_payment}} \\*
\quad GNN Miner          & 0   & 0.03258 & 18 & \phantom{0}3 & 3.432 & 0.191 \\*
                         & 0.5 & 0.10100 & 18 & \phantom{0}3 & 3.432 & 0.191 \\*
                         & 1   & 0.10352 & 18 & \phantom{0}3 & 3.432 & 0.191 \\*
                         & 10  & 0.10577 & 18 & \phantom{0}3 & 3.432 & 0.191 \\*
                         & inf & 0.10602 & 18 & \phantom{0}3 & 3.432 & 0.191 \\*
\addlinespace[3pt]
\quad Heuristics Miner   & 0   & 0.05268 & 18 & \phantom{0}3 & 4.878 & 0.271 \\*
                         & 0.5 & 0.15190 & 18 & \phantom{0}3 & 4.878 & 0.271 \\*
                         & 1   & 0.15941 & 18 & \phantom{0}3 & 4.878 & 0.271 \\*
                         & 10  & 0.16694 & 18 & \phantom{0}3 & 4.878 & 0.271 \\*
                         & inf & 0.16778 & 18 & \phantom{0}3 & 4.878 & 0.271 \\*
\addlinespace[3pt]
\quad ILP Miner          & 0   & 0.02308 & \phantom{0}7 & 14 & 3.524 & 0.503 \\*
                         & 0.5 & 0.15023 & \phantom{0}7 & 14 & 3.524 & 0.503 \\*
                         & 1   & 0.15325 & \phantom{0}7 & 14 & 3.524 & 0.503 \\*
                         & 10  & 0.15587 & \phantom{0}7 & 14 & 3.524 & 0.503 \\*
                         & inf & 0.15614 & \phantom{0}7 & 14 & 3.524 & 0.503 \\*
\addlinespace[3pt]
\quad Inductive Miner    & 0   & 0.02476 & 19 & \phantom{0}2 & 4.845 & 0.255 \\*
                         & 0.5 & 0.10289 & 19 & \phantom{0}2 & 4.845 & 0.255 \\*
                         & 1   & 0.10710 & 19 & \phantom{0}2 & 4.845 & 0.255 \\*
                         & 10  & 0.11100 & 19 & \phantom{0}2 & 4.845 & 0.255 \\*
                         & inf & 0.11144 & 19 & \phantom{0}2 & 4.845 & 0.255 \\*
\addlinespace[3pt]
\quad Split Miner        & 0   & 0.01701 & 20 & \phantom{0}1 & 4.024 & 0.201 \\*
                         & 0.5 & 0.07680 & 20 & \phantom{0}1 & 4.024 & 0.201 \\*
                         & 1   & 0.08202 & 20 & \phantom{0}1 & 4.024 & 0.201 \\*
                         & 10  & 0.08770 & 20 & \phantom{0}1 & 4.024 & 0.201 \\*
                         & inf & 0.08840 & 20 & \phantom{0}1 & 4.024 & 0.201 \\
\addlinespace[8pt]

\multicolumn{7}{@{}l}{\textbf{road\_traffic}} \\*
\quad GNN Miner          & 0   & 0.01657 & 13 & \phantom{0}0 & 5.319 & 0.409 \\*
                         & 0.5 & 0.09992 & 13 & \phantom{0}0 & 5.319 & 0.409 \\*
                         & 1   & 0.11121 & 13 & \phantom{0}0 & 5.319 & 0.409 \\*
                         & 10  & 0.12260 & 13 & \phantom{0}0 & 5.319 & 0.409 \\*
                         & inf & 0.12393 & 13 & \phantom{0}0 & 5.319 & 0.409 \\*
\addlinespace[3pt]
\quad Heuristics Miner   & 0   & 0.01334 & 13 & \phantom{0}0 & 3.345 & 0.257 \\*
                         & 0.5 & 0.05142 & 13 & \phantom{0}0 & 3.345 & 0.257 \\*
                         & 1   & 0.05383 & 13 & \phantom{0}0 & 3.345 & 0.257 \\*
                         & 10  & 0.05608 & 13 & \phantom{0}0 & 3.345 & 0.257 \\*
                         & inf & 0.05634 & 13 & \phantom{0}0 & 3.345 & 0.257 \\*
\addlinespace[3pt]
\quad ILP Miner          & 0   & 0.00986 & \phantom{0}7 & \phantom{0}6 & 3.712 & 0.530 \\*
                         & 0.5 & 0.21264 & \phantom{0}7 & \phantom{0}6 & 3.712 & 0.530 \\*
                         & 1   & 0.23243 & \phantom{0}7 & \phantom{0}6 & 3.712 & 0.530 \\*
                         & 10  & 0.24949 & \phantom{0}7 & \phantom{0}6 & 3.712 & 0.530 \\*
                         & inf & 0.25132 & \phantom{0}7 & \phantom{0}6 & 3.712 & 0.530 \\*
\addlinespace[3pt]
\quad Inductive Miner    & 0   & 0.03650 & 13 & \phantom{0}0 & 5.199 & 0.400 \\*
                         & 0.5 & 0.13549 & 13 & \phantom{0}0 & 5.199 & 0.400 \\*
                         & 1   & 0.14033 & 13 & \phantom{0}0 & 5.199 & 0.400 \\*
                         & 10  & 0.14465 & 13 & \phantom{0}0 & 5.199 & 0.400 \\*
                         & inf & 0.14512 & 13 & \phantom{0}0 & 5.199 & 0.400 \\*
\addlinespace[3pt]
\quad Split Miner        & 0   & 0.01306 & 13 & \phantom{0}0 & 3.685 & 0.283 \\*
                         & 0.5 & 0.06290 & 13 & \phantom{0}0 & 3.685 & 0.283 \\*
                         & 1   & 0.06403 & 13 & \phantom{0}0 & 3.685 & 0.283 \\*
                         & 10  & 0.06504 & 13 & \phantom{0}0 & 3.685 & 0.283 \\*
                         & inf & 0.06515 & 13 & \phantom{0}0 & 3.685 & 0.283 \\
\addlinespace[8pt]

\multicolumn{7}{@{}l}{\textbf{sepsis}} \\*
\quad GNN Miner          & 0   & 0.02475 & 16 & \phantom{0}2 & 6.508 & 0.407 \\*
                         & 0.5 & 0.16937 & 16 & \phantom{0}2 & 6.508 & 0.407 \\*
                         & 1   & 0.19511 & 16 & \phantom{0}2 & 6.508 & 0.407 \\*
                         & 10  & 0.22219 & 16 & \phantom{0}2 & 6.508 & 0.407 \\*
                         & inf & 0.22548 & 16 & \phantom{0}2 & 6.508 & 0.407 \\*
\addlinespace[3pt]
\quad Heuristics Miner   & 0   & 0.01138 & 18 & \phantom{0}0 & 3.303 & 0.184 \\*
                         & 0.5 & 0.09485 & 18 & \phantom{0}0 & 3.303 & 0.184 \\*
                         & 1   & 0.09730 & 18 & \phantom{0}0 & 3.303 & 0.184 \\*
                         & 10  & 0.09954 & 18 & \phantom{0}0 & 3.303 & 0.184 \\*
                         & inf & 0.09974 & 18 & \phantom{0}0 & 3.303 & 0.184 \\*
\addlinespace[3pt]
\quad ILP Miner          & 0   & 0.00664 & 13 & \phantom{0}5 & 3.554 & 0.273 \\*
                         & 0.5 & 0.08114 & 13 & \phantom{0}5 & 3.554 & 0.273 \\*
                         & 1   & 0.08584 & 13 & \phantom{0}5 & 3.554 & 0.273 \\*
                         & 10  & 0.09036 & 13 & \phantom{0}5 & 3.554 & 0.273 \\*
                         & inf & 0.09086 & 13 & \phantom{0}5 & 3.554 & 0.273 \\*
\addlinespace[3pt]
\quad Inductive Miner    & 0   & 0.04262 & 18 & \phantom{0}0 & 5.940 & 0.330 \\*
                         & 0.5 & 0.19027 & 18 & \phantom{0}0 & 5.940 & 0.330 \\*
                         & 1   & 0.19899 & 18 & \phantom{0}0 & 5.940 & 0.330 \\*
                         & 10  & 0.20734 & 18 & \phantom{0}0 & 5.940 & 0.330 \\*
                         & inf & 0.20830 & 18 & \phantom{0}0 & 5.940 & 0.330 \\*
\addlinespace[3pt]
\quad Split Miner        & 0   & 0.01072 & 18 & \phantom{0}0 & 4.527 & 0.251 \\*
                         & 0.5 & 0.07321 & 18 & \phantom{0}0 & 4.527 & 0.251 \\*
                         & 1   & 0.07596 & 18 & \phantom{0}0 & 4.527 & 0.251 \\*
                         & 10  & 0.07847 & 18 & \phantom{0}0 & 4.527 & 0.251 \\*
                         & inf & 0.07874 & 18 & \phantom{0}0 & 4.527 & 0.251 \\
\end{longtable}
\endgroup

Similar to the previous table, Tab. \ref{tab:dataset-method-sweeps-static} presents results under a static model scenario.

\begingroup
\small
\setlength{\tabcolsep}{0pt}
\setlength{\LTleft}{0pt}
\setlength{\LTright}{0pt plus 1fil}
\renewcommand{\arraystretch}{1.08}
\begin{longtable}{@{\extracolsep{\fill}}l c c c c c c@{}}
\caption{Static model distance of node embeddings, grouped by dataset and method, with varying $\lambda$.}
\label{tab:dataset-method-sweeps-static}\\
\toprule
Dataset / Method & $\lambda$ & $D_{\text{Weighted}}$ & \# Matched & \# Unmatched & Total residual & Per-anchor \\
\midrule
\endfirsthead
\multicolumn{7}{@{}l}{\textbf{\tablename~\thetable}\ (continued)}\\[2pt]
\toprule
Dataset / Method & $\lambda$ & $D_{\text{Weighted}}$ & \# Matched & \# Unmatched & Total residual & Per-anchor \\
\midrule
\endhead
\midrule
\multicolumn{7}{r@{}}{\footnotesize\itshape Continued on next page}\\
\endfoot
\bottomrule
\endlastfoot

\multicolumn{7}{@{}l}{\textbf{bpic\_2012}} \\*
\quad GNN Miner          & 0   & 0.00942 & 12 & 14 & 5.431 & 0.453 \\*
                         & 0.5 & 0.12431 & 12 & 14 & 5.431 & 0.453 \\*
                         & 1   & 0.12893 & 12 & 14 & 5.431 & 0.453 \\*
                         & 10  & 0.13306 & 12 & 14 & 5.431 & 0.453 \\*
                         & inf & 0.13351 & 12 & 14 & 5.431 & 0.453 \\*
\addlinespace[3pt]
\quad Heuristics Miner   & 0   & 0.00745 & 11 & 15 & 6.819 & 0.620 \\*
                         & 0.5 & 0.19744 & 11 & 15 & 6.819 & 0.620 \\*
                         & 1   & 0.22100 & 11 & 15 & 6.819 & 0.620 \\*
                         & 10  & 0.22273 & 11 & 15 & 6.819 & 0.620 \\*
                         & inf & 0.22787 & 11 & 15 & 6.819 & 0.620 \\*
\addlinespace[3pt]
\quad ILP Miner          & 0   & 0.04116 & \phantom{0}8 & 18 & 8.567 & 1.071 \\*
                         & 0.5 & 0.24079 & \phantom{0}8 & 18 & 8.567 & 1.071 \\*
                         & 1   & 0.25121 & \phantom{0}8 & 18 & 8.567 & 1.071 \\*
                         & 10  & 0.25985 & \phantom{0}8 & 18 & 8.567 & 1.071 \\*
                         & inf & 0.26074 & \phantom{0}8 & 18 & 8.567 & 1.071 \\*
\addlinespace[3pt]
\quad Inductive Miner    & 0   & 0.02657 & 11 & 15 & 6.612 & 0.601 \\*
                         & 0.5 & 0.27485 & 11 & 15 & 6.612 & 0.601 \\*
                         & 1   & 0.30744 & 11 & 15 & 6.612 & 0.601 \\*
                         & 10  & 0.33454 & 11 & 15 & 6.612 & 0.601 \\*
                         & inf & 0.33751 & 11 & 15 & 6.612 & 0.601 \\*
\addlinespace[3pt]
\quad Split Miner        & 0   & 0.01329 & 10 & 16 & 6.611 & 0.661 \\*
                         & 0.5 & 0.23556 & 10 & 16 & 6.611 & 0.661 \\*
                         & 1   & 0.24881 & 10 & 16 & 6.611 & 0.661 \\*
                         & 10  & 0.25934 & 10 & 16 & 6.611 & 0.661 \\*
                         & inf & 0.26044 & 10 & 16 & 6.611 & 0.661 \\
\addlinespace[8pt]

\multicolumn{7}{@{}l}{\textbf{bpic\_2017}} \\*
\quad GNN Miner          & 0   & 0.02255 & 10 & 18 & 5.724 & 0.572 \\*
                         & 0.5 & 0.17458 & 10 & 18 & 5.724 & 0.572 \\*
                         & 1   & 0.17354 & 10 & 18 & 5.724 & 0.572 \\*
                         & 10  & 0.17600 & 10 & 18 & 5.724 & 0.572 \\*
                         & inf & 0.17828 & 10 & 18 & 5.724 & 0.572 \\*
\addlinespace[3pt]
\quad Heuristics Miner   & 0   & 0.00265 & \phantom{0}9 & 19 & 4.614 & 0.513 \\*
                         & 0.5 & 0.19796 & \phantom{0}9 & 19 & 4.614 & 0.513 \\*
                         & 1   & 0.21102 & \phantom{0}9 & 19 & 4.614 & 0.513 \\*
                         & 10  & 0.22231 & \phantom{0}9 & 19 & 4.614 & 0.513 \\*
                         & inf & 0.22315 & \phantom{0}9 & 19 & 4.614 & 0.513 \\*
\addlinespace[3pt]
\quad ILP Miner          & 0   & 0.00734 & \phantom{0}6 & 22 & 4.105 & 0.684 \\*
                         & 0.5 & 0.23326 & \phantom{0}6 & 22 & 4.105 & 0.684 \\*
                         & 1   & 0.23967 & \phantom{0}6 & 22 & 4.105 & 0.684 \\*
                         & 10  & 0.24463 & \phantom{0}6 & 22 & 4.105 & 0.684 \\*
                         & inf & 0.24516 & \phantom{0}6 & 22 & 4.105 & 0.684 \\*
\addlinespace[3pt]
\quad Inductive Miner    & 0   & 0.01363 & \phantom{0}9 & 19 & 5.593 & 0.621 \\*
                         & 0.5 & 0.16545 & \phantom{0}9 & 19 & 5.593 & 0.621 \\*
                         & 1   & 0.17035 & \phantom{0}9 & 19 & 5.593 & 0.621 \\*
                         & 10  & 0.17460 & \phantom{0}9 & 19 & 5.593 & 0.621 \\*
                         & inf & 0.17504 & \phantom{0}9 & 19 & 5.593 & 0.621 \\*
\addlinespace[3pt]
\quad Split Miner        & 0   & 0.01018 & \phantom{0}8 & 20 & 3.586 & 0.448 \\*
                         & 0.5 & 0.17917 & \phantom{0}8 & 20 & 3.586 & 0.448 \\*
                         & 1   & 0.18381 & \phantom{0}8 & 20 & 3.586 & 0.448 \\*
                         & 10  & 0.18755 & \phantom{0}8 & 20 & 3.586 & 0.448 \\*
                         & inf & 0.18795 & \phantom{0}8 & 20 & 3.586 & 0.448 \\
\addlinespace[8pt]

\multicolumn{7}{@{}l}{\textbf{bpic\_2020\_Domestic\_declarations}} \\*
\quad GNN Miner          & 0   & 0.01549 & 19 & \phantom{0}0 & 8.304 & 0.437 \\*
                         & 0.5 & 0.15489 & 19 & \phantom{0}0 & 8.304 & 0.437 \\*
                         & 1   & 0.15812 & 19 & \phantom{0}0 & 8.304 & 0.437 \\*
                         & 10  & 0.16382 & 19 & \phantom{0}0 & 8.304 & 0.437 \\*
                         & inf & 0.16011 & 19 & \phantom{0}0 & 8.304 & 0.437 \\*
\addlinespace[3pt]
\quad Heuristics Miner   & 0   & 0.00146 & 18 & \phantom{0}1 & 5.905 & 0.328 \\*
                         & 0.5 & 0.11556 & 18 & \phantom{0}1 & 5.905 & 0.328 \\*
                         & 1   & 0.12618 & 18 & \phantom{0}1 & 5.905 & 0.328 \\*
                         & 10  & 0.13752 & 18 & \phantom{0}1 & 5.905 & 0.328 \\*
                         & inf & 0.13889 & 18 & \phantom{0}1 & 5.905 & 0.328 \\*
\addlinespace[3pt]
\quad ILP Miner          & 0   & 0.00601 & \phantom{0}6 & 13 & 3.842 & 0.640 \\*
                         & 0.5 & 0.16950 & \phantom{0}6 & 13 & 3.842 & 0.640 \\*
                         & 1   & 0.17394 & \phantom{0}6 & 13 & 3.842 & 0.640 \\*
                         & 10  & 0.17754 & \phantom{0}6 & 13 & 3.842 & 0.640 \\*
                         & inf & 0.17792 & \phantom{0}6 & 13 & 3.842 & 0.640 \\*
\addlinespace[3pt]
\quad Inductive Miner    & 0   & 0.03118 & 18 & \phantom{0}1 & 6.690 & 0.372 \\*
                         & 0.5 & 0.19046 & 18 & \phantom{0}1 & 6.690 & 0.372 \\*
                         & 1   & 0.21903 & 18 & \phantom{0}1 & 6.690 & 0.372 \\*
                         & 10  & 0.24771 & 18 & \phantom{0}1 & 6.690 & 0.372 \\*
                         & inf & 0.25100 & 18 & \phantom{0}1 & 6.690 & 0.372 \\*
\addlinespace[3pt]
\quad Split Miner        & 0   & 0.01361 & 17 & \phantom{0}2 & 6.191 & 0.364 \\*
                         & 0.5 & 0.14366 & 17 & \phantom{0}2 & 6.191 & 0.364 \\*
                         & 1   & 0.14659 & 17 & \phantom{0}2 & 6.191 & 0.364 \\*
                         & 10  & 0.14927 & 17 & \phantom{0}2 & 6.191 & 0.364 \\*
                         & inf & 0.15076 & 17 & \phantom{0}2 & 6.191 & 0.364 \\
\addlinespace[8pt]

\multicolumn{7}{@{}l}{\textbf{bpic\_2020\_International\_declarations}} \\*
\quad GNN Miner          & 0   & 0.02871 & 20 & 16 & 7.030 & 0.351 \\*
                         & 0.5 & 0.17710 & 20 & 16 & 7.030 & 0.351 \\*
                         & 1   & 0.18322 & 20 & 16 & 7.030 & 0.351 \\*
                         & 10  & 0.18884 & 20 & 16 & 7.030 & 0.351 \\*
                         & inf & 0.18945 & 20 & 16 & 7.030 & 0.351 \\*
\addlinespace[3pt]
\quad Heuristics Miner   & 0   & 0.03019 & 19 & 17 & 5.171 & 0.272 \\*
                         & 0.5 & 0.15752 & 19 & 17 & 5.171 & 0.272 \\*
                         & 1   & 0.16631 & 19 & 17 & 5.171 & 0.272 \\*
                         & 10  & 0.17454 & 19 & 17 & 5.171 & 0.272 \\*
                         & inf & 0.17524 & 19 & 17 & 5.171 & 0.272 \\*
\addlinespace[3pt]
\quad ILP Miner          & 0   & 0.01984 & 11 & 25 & 5.229 & 0.475 \\*
                         & 0.5 & 0.15468 & 11 & 25 & 5.229 & 0.475 \\*
                         & 1   & 0.15718 & 11 & 25 & 5.229 & 0.475 \\*
                         & 10  & 0.15947 & 11 & 25 & 5.229 & 0.475 \\*
                         & inf & 0.15971 & 11 & 25 & 5.229 & 0.475 \\*
\addlinespace[3pt]
\quad Inductive Miner    & 0   & 0.06760 & 19 & 17 & 8.325 & 0.438 \\*
                         & 0.5 & 0.35280 & 19 & 17 & 8.325 & 0.438 \\*
                         & 1   & 0.36996 & 19 & 17 & 8.325 & 0.438 \\*
                         & 10  & 0.38363 & 19 & 17 & 8.325 & 0.438 \\*
                         & inf & 0.38505 & 19 & 17 & 8.325 & 0.438 \\*
\addlinespace[3pt]
\quad Split Miner        & 0   & 0.04174 & 18 & 18 & 6.461 & 0.359 \\*
                         & 0.5 & 0.14492 & 18 & 18 & 6.461 & 0.359 \\*
                         & 1   & 0.15359 & 18 & 18 & 6.461 & 0.359 \\*
                         & 10  & 0.15037 & 18 & 18 & 6.461 & 0.359 \\*
                         & inf & 0.15221 & 18 & 18 & 6.461 & 0.359 \\
\addlinespace[8pt]

\multicolumn{7}{@{}l}{\textbf{bpic\_2020\_Permit\_log}} \\*
\quad GNN Miner          & 0   & 0.07706 & 20 & 33 & 6.946 & 0.347 \\*
                         & 0.5 & 0.33126 & 20 & 33 & 6.946 & 0.347 \\*
                         & 1   & 0.33207 & 20 & 33 & 6.946 & 0.347 \\*
                         & 10  & 0.33944 & 20 & 33 & 6.946 & 0.347 \\*
                         & inf & 0.34127 & 20 & 33 & 6.946 & 0.347 \\*
\addlinespace[3pt]
\quad Heuristics Miner   & 0   & 0.02823 & 19 & 34 & 5.070 & 0.267 \\*
                         & 0.5 & 0.16646 & 19 & 34 & 5.070 & 0.267 \\*
                         & 1   & 0.16958 & 19 & 34 & 5.070 & 0.267 \\*
                         & 10  & 0.17245 & 19 & 34 & 5.070 & 0.267 \\*
                         & inf & 0.17274 & 19 & 34 & 5.070 & 0.267 \\*
\addlinespace[3pt]
\quad ILP Miner          & 0   & 0.03397 & 11 & 42 & 4.509 & 0.410 \\*
                         & 0.5 & 0.20057 & 11 & 42 & 4.509 & 0.410 \\*
                         & 1   & 0.20506 & 11 & 42 & 4.509 & 0.410 \\*
                         & 10  & 0.20912 & 11 & 42 & 4.509 & 0.410 \\*
                         & inf & 0.20953 & 11 & 42 & 4.509 & 0.410 \\*
\addlinespace[3pt]
\quad Inductive Miner    & 0   & 0.04455 & 19 & 34 & 8.083 & 0.425 \\*
                         & 0.5 & 0.41445 & 19 & 34 & 8.083 & 0.425 \\*
                         & 1   & 0.42508 & 19 & 34 & 8.083 & 0.425 \\*
                         & 10  & 0.43311 & 19 & 34 & 8.083 & 0.425 \\*
                         & inf & 0.43394 & 19 & 34 & 8.083 & 0.425 \\*
\addlinespace[3pt]
\quad Split Miner        & 0   & 0.03111 & 18 & 35 & 6.257 & 0.348 \\*
                         & 0.5 & 0.14612 & 18 & 35 & 6.257 & 0.348 \\*
                         & 1   & 0.15001 & 18 & 35 & 6.257 & 0.348 \\*
                         & 10  & 0.15073 & 18 & 35 & 6.257 & 0.348 \\*
                         & inf & 0.15144 & 18 & 35 & 6.257 & 0.348 \\
\addlinespace[8pt]

\multicolumn{7}{@{}l}{\textbf{bpic\_2020\_Prepaid\_travel\_cost}} \\*
\quad GNN Miner          & 0   & 0.06543 & 20 & 11 & 2.265 & 0.113 \\*
                         & 0.5 & 0.21660 & 20 & 11 & 2.265 & 0.113 \\*
                         & 1   & 0.22713 & 20 & 11 & 2.265 & 0.113 \\*
                         & 10  & 0.23683 & 20 & 11 & 2.265 & 0.113 \\*
                         & inf & 0.23792 & 20 & 11 & 2.265 & 0.113 \\*
\addlinespace[3pt]
\quad Heuristics Miner   & 0   & 0.03584 & 19 & 12 & 6.088 & 0.320 \\*
                         & 0.5 & 0.13510 & 19 & 12 & 6.088 & 0.320 \\*
                         & 1   & 0.14470 & 19 & 12 & 6.088 & 0.320 \\*
                         & 10  & 0.15354 & 19 & 12 & 6.088 & 0.320 \\*
                         & inf & 0.15466 & 19 & 12 & 6.088 & 0.320 \\*
\addlinespace[3pt]
\quad ILP Miner          & 0   & 0.00919 & 11 & 20 & 5.564 & 0.506 \\*
                         & 0.5 & 0.15083 & 11 & 20 & 5.564 & 0.506 \\*
                         & 1   & 0.15362 & 11 & 20 & 5.564 & 0.506 \\*
                         & 10  & 0.15608 & 11 & 20 & 5.564 & 0.506 \\*
                         & inf & 0.15633 & 11 & 20 & 5.564 & 0.506 \\*
\addlinespace[3pt]
\quad Inductive Miner    & 0   & 0.01781 & 19 & 12 & 9.509 & 0.500 \\*
                         & 0.5 & 0.42579 & 19 & 12 & 9.509 & 0.500 \\*
                         & 1   & 0.51100 & 19 & 12 & 9.509 & 0.500 \\*
                         & 10  & 0.57938 & 19 & 12 & 9.509 & 0.500 \\*
                         & inf & 0.58553 & 19 & 12 & 9.509 & 0.500 \\*
\addlinespace[3pt]
\quad Split Miner        & 0   & 0.04383 & 18 & 13 & 7.093 & 0.394 \\*
                         & 0.5 & 0.12859 & 18 & 13 & 7.093 & 0.394 \\*
                         & 1   & 0.12975 & 18 & 13 & 7.093 & 0.394 \\*
                         & 10  & 0.13876 & 18 & 13 & 7.093 & 0.394 \\*
                         & inf & 0.14100 & 18 & 13 & 7.093 & 0.394 \\
\addlinespace[8pt]

\multicolumn{7}{@{}l}{\textbf{bpic\_2020\_Request\_for\_payment}} \\*
\quad GNN Miner          & 0   & 0.01429 & 20 & \phantom{0}1 & 6.204 & 0.310 \\*
                         & 0.5 & 0.15507 & 20 & \phantom{0}1 & 6.204 & 0.310 \\*
                         & 1   & 0.18176 & 20 & \phantom{0}1 & 6.204 & 0.310 \\*
                         & 10  & 0.21891 & 20 & \phantom{0}1 & 6.204 & 0.310 \\*
                         & inf & 0.22406 & 20 & \phantom{0}1 & 6.204 & 0.310 \\*
\addlinespace[3pt]
\quad Heuristics Miner   & 0   & 0.02907 & 19 & \phantom{0}2 & 6.192 & 0.326 \\*
                         & 0.5 & 0.16051 & 19 & \phantom{0}2 & 6.192 & 0.326 \\*
                         & 1   & 0.17232 & 19 & \phantom{0}2 & 6.192 & 0.326 \\*
                         & 10  & 0.18467 & 19 & \phantom{0}2 & 6.192 & 0.326 \\*
                         & inf & 0.18615 & 19 & \phantom{0}2 & 6.192 & 0.326 \\*
\addlinespace[3pt]
\quad ILP Miner          & 0   & 0.00835 & \phantom{0}6 & 15 & 3.710 & 0.618 \\*
                         & 0.5 & 0.11685 & \phantom{0}6 & 15 & 3.710 & 0.618 \\*
                         & 1   & 0.11769 & \phantom{0}6 & 15 & 3.710 & 0.618 \\*
                         & 10  & 0.11840 & \phantom{0}6 & 15 & 3.710 & 0.618 \\*
                         & inf & 0.11848 & \phantom{0}6 & 15 & 3.710 & 0.618 \\*
\addlinespace[3pt]
\quad Inductive Miner    & 0   & 0.02225 & 19 & \phantom{0}2 & 9.487 & 0.499 \\*
                         & 0.5 & 0.18768 & 19 & \phantom{0}2 & 9.487 & 0.499 \\*
                         & 1   & 0.20095 & 19 & \phantom{0}2 & 9.487 & 0.499 \\*
                         & 10  & 0.21321 & 19 & \phantom{0}2 & 9.487 & 0.499 \\*
                         & inf & 0.21452 & 19 & \phantom{0}2 & 9.487 & 0.499 \\*
\addlinespace[3pt]
\quad Split Miner        & 0   & 0.05418 & 18 & \phantom{0}3 & 7.037 & 0.391 \\*
                         & 0.5 & 0.10335 & 18 & \phantom{0}3 & 7.037 & 0.391 \\*
                         & 1   & 0.10353 & 18 & \phantom{0}3 & 7.037 & 0.391 \\*
                         & 10  & 0.10558 & 18 & \phantom{0}3 & 7.037 & 0.391 \\*
                         & inf & 0.10837 & 18 & \phantom{0}3 & 7.037 & 0.391 \\
\addlinespace[8pt]

\multicolumn{7}{@{}l}{\textbf{road\_traffic}} \\*
\quad GNN Miner          & 0   & 0.04270 & 13 & \phantom{0}0 & 5.363 & 0.413 \\*
                         & 0.5 & 0.10830 & 13 & \phantom{0}0 & 5.363 & 0.413 \\*
                         & 1   & 0.11555 & 13 & \phantom{0}0 & 5.363 & 0.413 \\*
                         & 10  & 0.12321 & 13 & \phantom{0}0 & 5.363 & 0.413 \\*
                         & inf & 0.12411 & 13 & \phantom{0}0 & 5.363 & 0.413 \\*
\addlinespace[3pt]
\quad Heuristics Miner   & 0   & 0.01255 & 12 & \phantom{0}1 & 3.654 & 0.305 \\*
                         & 0.5 & 0.07203 & 12 & \phantom{0}1 & 3.654 & 0.305 \\*
                         & 1   & 0.08630 & 12 & \phantom{0}1 & 3.654 & 0.305 \\*
                         & 10  & 0.10338 & 12 & \phantom{0}1 & 3.654 & 0.305 \\*
                         & inf & 0.10558 & 12 & \phantom{0}1 & 3.654 & 0.305 \\*
\addlinespace[3pt]
\quad ILP Miner          & 0   & 0.03778 & \phantom{0}6 & \phantom{0}7 & 4.137 & 0.689 \\*
                         & 0.5 & 0.17195 & \phantom{0}6 & \phantom{0}7 & 4.137 & 0.689 \\*
                         & 1   & 0.17309 & \phantom{0}6 & \phantom{0}7 & 4.137 & 0.689 \\*
                         & 10  & 0.18210 & \phantom{0}6 & \phantom{0}7 & 4.137 & 0.689 \\*
                         & inf & 0.18309 & \phantom{0}6 & \phantom{0}7 & 4.137 & 0.689 \\*
\addlinespace[3pt]
\quad Inductive Miner    & 0   & 0.01149 & 12 & \phantom{0}1 & 4.946 & 0.412 \\*
                         & 0.5 & 0.14471 & 12 & \phantom{0}1 & 4.946 & 0.412 \\*
                         & 1   & 0.16573 & 12 & \phantom{0}1 & 4.946 & 0.412 \\*
                         & 10  & 0.19060 & 12 & \phantom{0}1 & 4.946 & 0.412 \\*
                         & inf & 0.19375 & 12 & \phantom{0}1 & 4.946 & 0.412 \\*
\addlinespace[3pt]
\quad Split Miner        & 0   & 0.02501 & 11 & \phantom{0}2 & 3.668 & 0.333 \\*
                         & 0.5 & 0.11815 & 11 & \phantom{0}2 & 3.668 & 0.333 \\*
                         & 1   & 0.12541 & 11 & \phantom{0}2 & 3.668 & 0.333 \\*
                         & 10  & 0.13230 & 11 & \phantom{0}2 & 3.668 & 0.333 \\*
                         & inf & 0.13308 & 11 & \phantom{0}2 & 3.668 & 0.333 \\
\addlinespace[8pt]

\multicolumn{7}{@{}l}{\textbf{sepsis}} \\*
\quad GNN Miner          & 0   & 0.03623 & 18 & \phantom{0}0 & 4.741 & 0.263 \\*
                         & 0.5 & 0.14301 & 18 & \phantom{0}0 & 4.741 & 0.263 \\*
                         & 1   & 0.16655 & 18 & \phantom{0}0 & 4.741 & 0.263 \\*
                         & 10  & 0.19545 & 18 & \phantom{0}0 & 4.741 & 0.263 \\*
                         & inf & 0.19924 & 18 & \phantom{0}0 & 4.741 & 0.263 \\*
\addlinespace[3pt]
\quad Heuristics Miner   & 0   & 0.03000 & 17 & \phantom{0}1 & 3.048 & 0.179 \\*
                         & 0.5 & 0.10353 & 17 & \phantom{0}1 & 3.048 & 0.179 \\*
                         & 1   & 0.11193 & 17 & \phantom{0}1 & 3.048 & 0.179 \\*
                         & 10  & 0.12041 & 17 & \phantom{0}1 & 3.048 & 0.179 \\*
                         & inf & 0.12139 & 17 & \phantom{0}1 & 3.048 & 0.179 \\*
\addlinespace[3pt]
\quad ILP Miner          & 0   & 0.03587 & 12 & \phantom{0}6 & 5.528 & 0.461 \\*
                         & 0.5 & 0.12785 & 12 & \phantom{0}6 & 5.528 & 0.461 \\*
                         & 1   & 0.13531 & 12 & \phantom{0}6 & 5.528 & 0.461 \\*
                         & 10  & 0.14217 & 12 & \phantom{0}6 & 5.528 & 0.461 \\*
                         & inf & 0.14294 & 12 & \phantom{0}6 & 5.528 & 0.461 \\*
\addlinespace[3pt]
\quad Inductive Miner    & 0   & 0.02740 & 17 & \phantom{0}1 & 4.161 & 0.245 \\*
                         & 0.5 & 0.17099 & 17 & \phantom{0}1 & 4.161 & 0.245 \\*
                         & 1   & 0.18557 & 17 & \phantom{0}1 & 4.161 & 0.245 \\*
                         & 10  & 0.20058 & 17 & \phantom{0}1 & 4.161 & 0.245 \\*
                         & inf & 0.20237 & 17 & \phantom{0}1 & 4.161 & 0.245 \\*
\addlinespace[3pt]
\quad Split Miner        & 0   & 0.06435 & 16 & \phantom{0}2 & 4.029 & 0.252 \\*
                         & 0.5 & 0.10046 & 16 & \phantom{0}2 & 4.029 & 0.252 \\*
                         & 1   & 0.10355 & 16 & \phantom{0}2 & 4.029 & 0.252 \\*
                         & 10  & 0.10366 & 16 & \phantom{0}2 & 4.029 & 0.252 \\*
                         & inf & 0.10476 & 16 & \phantom{0}2 & 4.029 & 0.252 \\
\end{longtable}
\endgroup

\section{Context-Aware and Static Fitness and Precision vs Node Embedding Distance}
\label{comp2}
\begin{figure}[!htbp]
    \centering
    \includegraphics[width=0.84\linewidth]{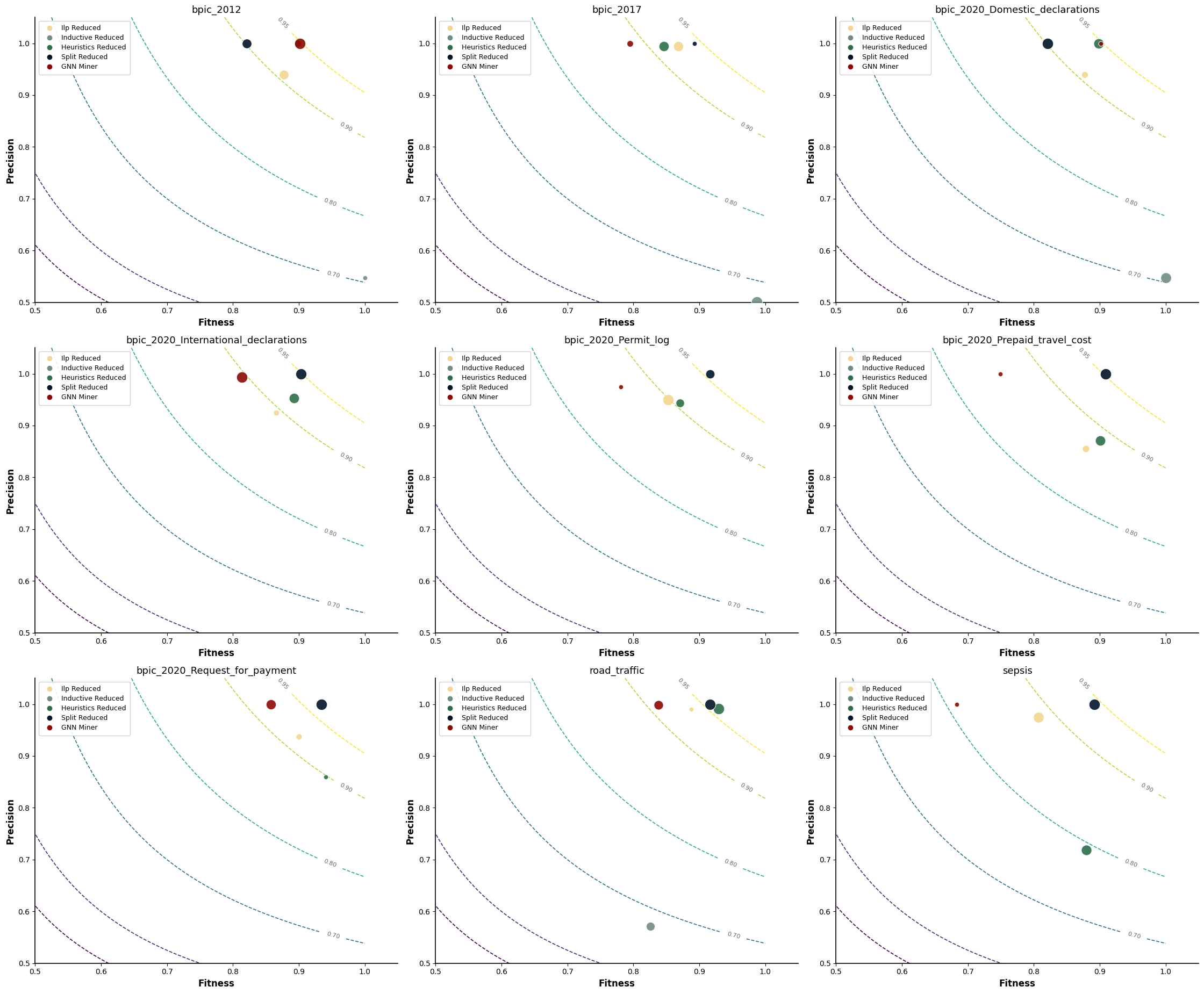}
    \caption{Context-aware process model conformance checking results vs. cosine distance}
    \label{fig:appendixcontext}
\end{figure}

\begin{figure}[!htbp]
    \centering
    \includegraphics[width=0.84\linewidth]{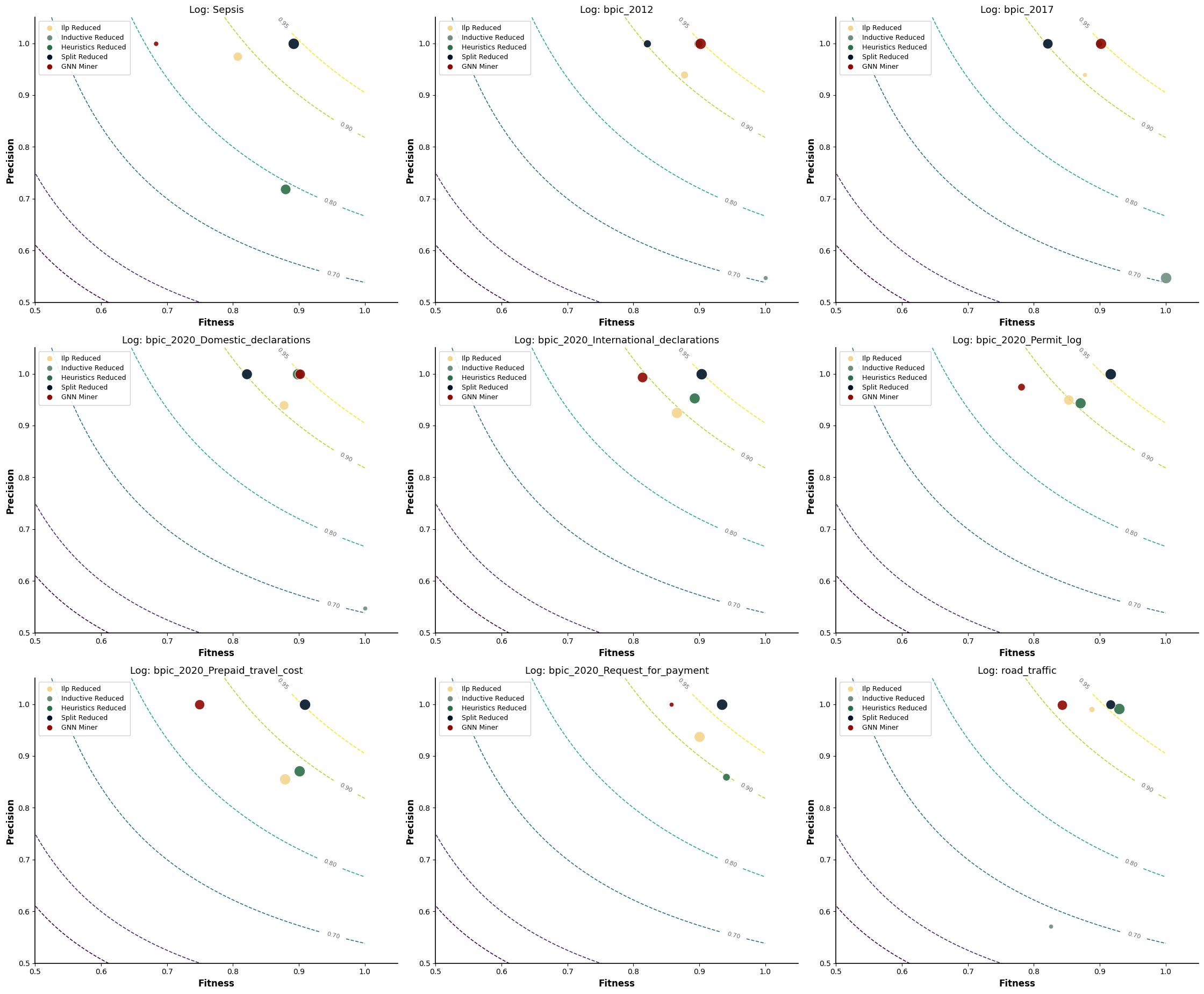}
    \caption{Static process model conformance checking results vs. cosine distance}
    \label{fig:appendixstatic}
\end{figure}

Fig.~\ref{fig:appendixcontext} and Fig.~\ref{fig:appendixstatic} illustrate the relationship between fitness (x-axis), precision (y-axis), and the normalized distance values represented by point sizes. Each point’s size corresponds to the similarity between the ground-truth and the discovered process model, where larger points indicate higher similarity. For readability, both axes are bounded between 0.5 and 1, and low-performing methods are excluded. As shown, aside from a few exceptions discussed in Section~\ref{stcoresult}, an upward trend toward higher fitness and precision generally corresponds to larger point sizes, reflecting stronger alignment between the models and their respective event logs.

\clearpage
\section{Parameter Sensitivity Analysis}
\label{paramanalysis}
To assess the robustness of the proposed framework with respect to major hyperparameters, we report a sensitivity analysis over embedding dimension \(d\), penalty factor \(\alpha\), and minimum count.

To analyze the effect of embedding dimension \(d\) on the total Procrustes alignment residual, we include \texttt{BPIC\_2020\_Permit\_log} with ILP Miner as a representative challenging case, since it exhibits one of the smallest shared anchor sets among the evaluated settings. The results show that, although the residual values vary across embedding dimensions, the overall pattern remains comparable across settings.

\begin{table}[!htb]
\centering
\caption{Parameter sensitivity analysis on \texttt{BPIC\_2020\_Permit\_log} using ILP Miner, reported for different embedding dimensions \(d\).}
\label{tab:appendix_procrustes_dim_analysis}
\setlength{\tabcolsep}{8pt}
\begin{tabular}{@{}lrcccc@{}}
\toprule
Scenario & \(d\) & \# Matched & \# Unmatched & Total residual & Per-anchor residual \\
\midrule
context\_aware & 12  & 12 & 41 & 4.843588 & 0.403632 \\
context\_aware & 32  & 12 & 41 & 4.048280 & 0.337357 \\
context\_aware & 64  & 12 & 41 & 4.909243 & 0.409104 \\
context\_aware & 128 & 12 & 41 & 4.534481 & 0.377873 \\
\midrule
static        & 12  & 11 & 42 & 4.653875 & 0.423080 \\
static        & 32  & 11 & 42 & 4.713240 & 0.428476 \\
static        & 64  & 11 & 42 & 4.254982 & 0.386817 \\
static        & 128 & 11 & 42 & 4.508801 & 0.409891 \\
\bottomrule
\end{tabular}
\end{table}

We additionally examine the sensitivity of the framework to the unmatched-node penalty parameter \(\alpha\), together with the aggregation parameter \(\lambda\). For this analysis, we use the Sepsis dataset and vary \(\alpha\) and \(\lambda\) while holding the remaining parameters fixed. The purpose of this ablation is to make explicit how the final score depends on the treatment of unmatched nodes and on the degree of emphasis placed on larger node-level discrepancies.
The roles of \(\alpha\) and \(\lambda\) should be distinguished clearly. The parameter \(\alpha\) affects the magnitude of the penalty assigned specifically to unmatched nodes, whereas \(\lambda\) controls how strongly larger node-level distances influence the final aggregation more broadly. The two parameters therefore affect \(D_{\text{weighted}}\) in related but distinct ways.
Tab.~\ref{tab:sensitivity_context} and Tab.~\ref{tab:sensitivity_static} show the joint effect of \(\alpha\) and \(\lambda\) on the average distance for the context-aware and static settings, respectively. Across both tables, the general tendency is that larger \(\alpha\) and larger \(\lambda\) produce higher distances, with the largest values occurring when both parameters are high. This is consistent with the intended interpretation of the framework: stronger unmatched-node penalties and stronger emphasis on large discrepancies both increase the final conformance distance. At the same time, the context-aware setting remains lower than the static setting across most of the reported parameter grid (22 of 25 settings), indicating that this comparative advantage is not tied to a single \((\lambda,\alpha)\) choice.

\begin{table}[!htb]
\setlength{\tabcolsep}{5pt}
\begin{minipage}[t]{0.48\textwidth}
\centering
\caption{Sensitivity of context-aware average distance (Sepsis) to $\lambda$ and $\alpha$.}
\label{tab:sensitivity_context}
\begin{tabular}{@{}lccccc@{}}
\toprule
& \multicolumn{5}{c}{$\alpha$} \\
\cmidrule(l){2-6}
$\lambda$ & 0 & 0.25 & 0.5 & 0.75 & 1 \\
\midrule
0    & 0.0000 & 0.0060 & 0.0073 & 0.0079 & 0.0123 \\
0.5  & 0.0064 & 0.0097 & 0.0285 & 0.0391 & 0.0474 \\
1    & 0.0074 & 0.0093 & 0.0217 & 0.0711 & 0.0891 \\
10   & 0.0082 & 0.0307 & 0.0317 & 0.1092 & 0.0946 \\
$\infty$ & 0.0185 & 0.0258 & 0.0338 & 0.1088 & 0.1406 \\
\bottomrule
\end{tabular}
\end{minipage}%
\hfill
\begin{minipage}[t]{0.48\textwidth}
\centering
\caption{Sensitivity of static average distance (Sepsis) to $\lambda$ and $\alpha$.}
\label{tab:sensitivity_static}
\begin{tabular}{@{}lccccc@{}}
\toprule
& \multicolumn{5}{c}{$\alpha$} \\
\cmidrule(l){2-6}
$\lambda$ & 0 & 0.25 & 0.5 & 0.75 & 1 \\
\midrule
0    & 0.0000 & 0.0353 & 0.0324 & 0.0392 & 0.0384 \\
0.5  & 0.0264 & 0.0483 & 0.0576 & 0.0891 & 0.0819 \\
1    & 0.0326 & 0.0581 & 0.0694 & 0.0865 & 0.1024 \\
10   & 0.0632 & 0.0683 & 0.0943 & 0.1059 & 0.1272 \\
$\infty$ & 0.0527 & 0.0691 & 0.0704 & 0.1007 & 0.1541 \\
\bottomrule
\end{tabular}
\end{minipage}
\end{table}

To assess whether the Word2Vec \texttt{min\_count} parameter materially affects the embedding corpus, we varied \texttt{min\_count} over \(\{20, 30, 50, 100, 150, 200\}\) on the Sepsis dataset while keeping the remaining parameters fixed. Tab.~\ref{tab:mincount} reports the resulting number of generated training pairs for the trace-based and ground-truth embeddings under both the context-aware and static settings. Across all tested thresholds, the pair counts remain highly stable, with variations below \(1\%\) in all types of pairs. This indicates that, under the recommended walk-generation configuration, \texttt{min\_count} does not have a material effect on the effective embedding corpus. In this sense, \texttt{min\_count} acts primarily as a safety guard rather than as an active filter. We note, however, that in settings with substantially fewer walks or much rarer node occurrences, \texttt{min\_count} could become more influential; the present analysis is intended to document its behavior under the configuration used in this study.

\begin{table}[!htb]
\centering
\caption{Average number of Word2Vec embedding pairs across \texttt{min\_count} values (Sepsis).}
\label{tab:mincount}
\setlength{\tabcolsep}{8pt}
\begin{tabular}{@{}rcccc@{}}
\toprule
& \multicolumn{2}{c}{Context-aware} & \multicolumn{2}{c}{Static} \\
\cmidrule(lr){2-3}\cmidrule(l){4-5}
\texttt{min\_count} & Trace pairs & Truth pairs & Trace pairs & Truth pairs \\
\midrule
20  & 1,365,674 & 1,944,968 & 2,527,037 & 2,427,754 \\
30  & 1,368,732 & 1,943,188 & 2,530,069 & 2,427,626 \\
50  & 1,364,554 & 1,946,840 & 2,524,866 & 2,428,312 \\
100 & 1,356,886 & 1,940,308 & 2,527,260 & 2,430,278 \\
150 & 1,363,682 & 1,939,820 & 2,528,345 & 2,427,590 \\
200 & 1,352,494 & 1,920,433 & 2,521,543 & 2,421,120 \\
\bottomrule
\end{tabular}
\end{table}

The small fluctuations reflect the inherent stochasticity of the random walk process rather than systematic vocabulary filtering. This holds for both context-aware and static embeddings. Because random walk generation is computationally negligible (Appendix~\ref{runtime}), we recommend generating a sufficiently large walk corpus so that \texttt{min\_count} serves only as a safety guard, ensuring that the set of unmatched nodes is determined entirely by the discovery algorithm rather than by an embedding-level artifact.

\section{Runtime Analysis}
\label{runtime}
To assess the practical computational profile of the proposed framework, we report wall-clock runtime measurements across all nine benchmark datasets for both the context-aware and static scenarios. The analysis decomposes the pipeline into graph construction, random walk generation, Word2Vec training, and the final Procrustes alignment plus distance computation.

\begin{figure}[!htbp]
    \centering
    \includegraphics[width=\linewidth]{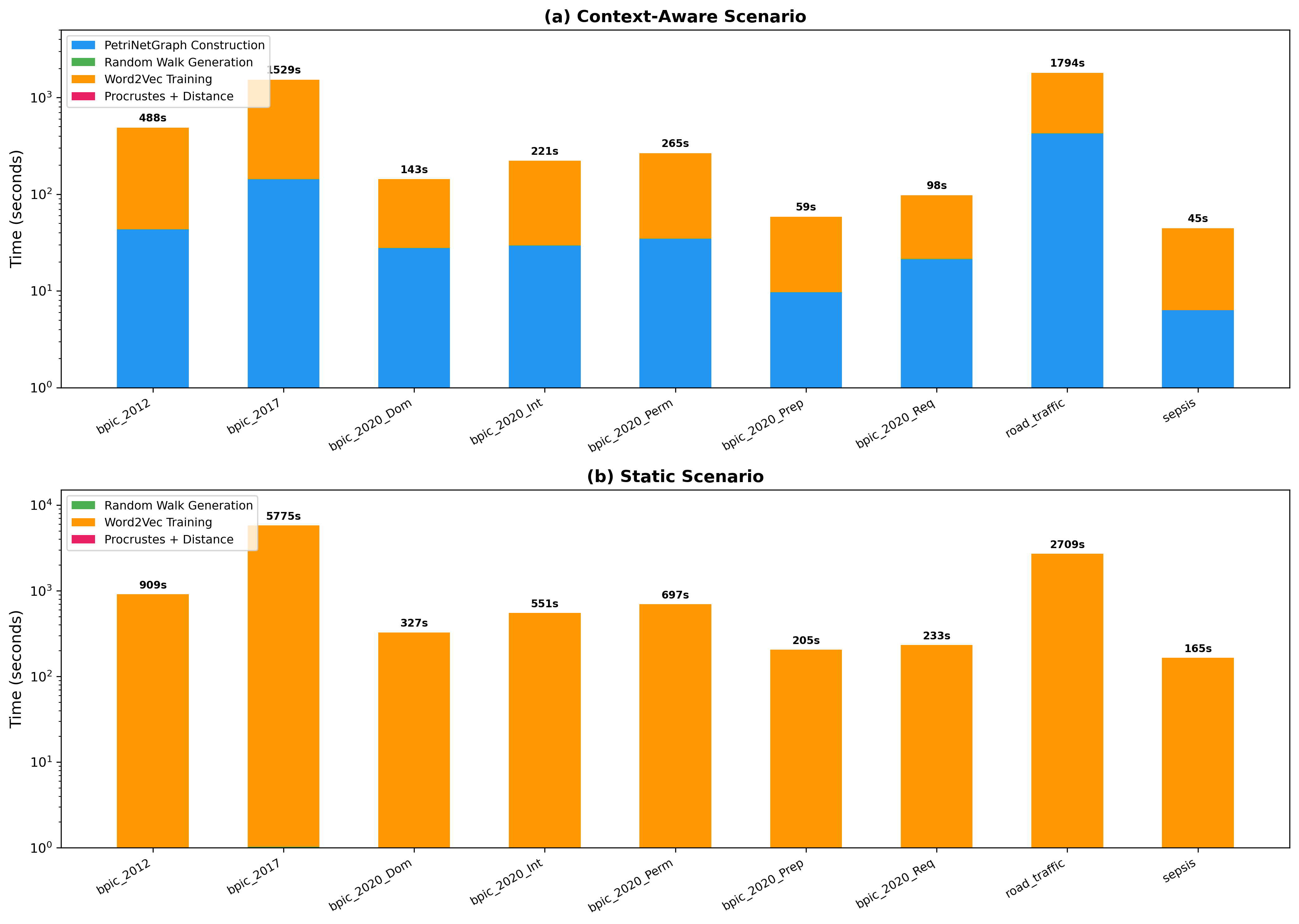}
    \caption{Wall-clock runtime breakdown across the nine benchmark datasets for the two evaluation scenarios}
    \label{fig:runtime_breakdown}
\end{figure}

Fig.~\ref{fig:runtime_breakdown} shows that Word2Vec training is the dominant cost in both scenarios. In the context-aware setting, the graph-construction step is the second main contributor, since a separate graph must be constructed for each trace. In the static setting, graph construction is performed only once and is therefore negligible in comparison with the embedding stage. Across all datasets, the Procrustes alignment and final distance computation contribute only a very small fraction of the total runtime. The total runtime nevertheless differs between the two scenarios. In the context-aware setting, total runtime ranges from 45\,s on sepsis to 1{,}794\,s on road traffic dataset. In the static setting, total runtime ranges from 165\,s to 5{,}775\,s. Although the static scenario avoids repeated graph construction, it can be slower overall because it generates more Word2Vec training pairs, which increases the embedding time substantially.

\begin{figure}[!htbp]
    \centering
    \includegraphics[width=\linewidth]{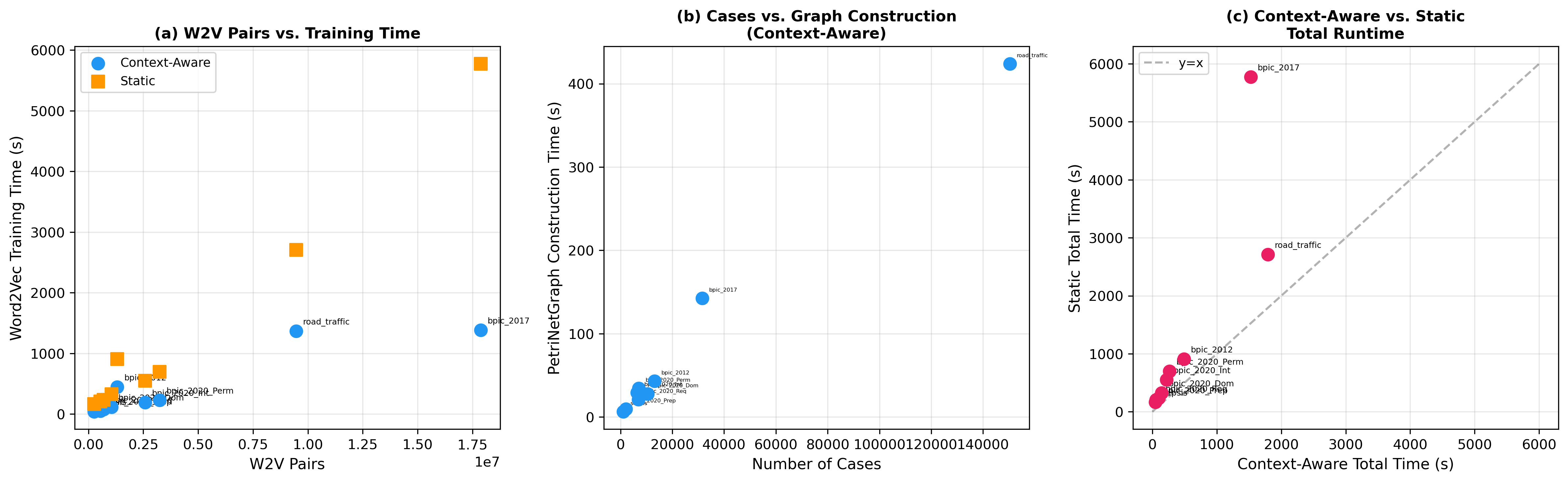}
    \caption{Runtime scaling behavior of the proposed framework}
    \label{fig:runtime_scaling}
\end{figure}

Fig.~\ref{fig:runtime_scaling} further clarifies the scaling behavior of the framework. Panel~(a) shows the relation between Word2Vec training time and the number of generated training pairs in both scenarios. For direct comparability, the static scenario is reported on the same normalized basis as the corresponding context-aware dataset counterpart, by proportionally rescaling both the training-pair counts and the associated training times. This allows the plot to emphasize the relative scaling trend of the embedding stage across scenarios rather than absolute volume differences. Panel~(b) shows that, in the context-aware setting, graph-construction time increases with the number of cases. Panel~(c) compares total runtime between the two scenarios and shows that the static setting is often slower overall, despite its lower graph-construction cost. Overall, these results indicate that the dominant computational cost lies in the embedding stage rather than in the alignment and aggregation components introduced by the proposed framework.

\end{document}